\documentclass[twocolumn]{settingfiles/autart}    

\usepackage{graphicx}
\usepackage{color}
\usepackage{amsfonts}
\usepackage{amsmath}
\usepackage{amssymb}
\usepackage{enumerate}
\usepackage{url}
\usepackage[overload]{empheq}
\usepackage{marginnote}

\begin{document}

\begin{frontmatter}

	\title{Control of a Quadrotor and a Ground Vehicle Manipulating an Object\thanksref{footnoteinfo}} 

	\thanks[footnoteinfo]{This work is supported by FRIA "COPTERS", Ref. F-3-5-5/FRIA/FC-12687. The preliminary results of this paper were presented at the IEEE American Control Conference (ACC), 2016 (see \cite{nguyen2016control}).}

	\author[ULB]{Tam W. Nguyen}\ead{tanguyen@ulb.ac.be}, 
	\author[ULB]{Laurent Catoire}\ead{laurent.catoire@ulb.ac.be},
	\author[ULB]{Emanuele Garone}\ead{egarone@ulb.ac.be} 

	\address[ULB]{Universit\'{e} libre de Bruxelles, Av. F. Roosevelt 50, CP 165/55, 1050, Brussels, Belgium} 

	\begin{abstract}
		This paper focuses on the control of a cooperative system composed of an Unmanned Aerial Vehicle (UAV) and an Unmanned Ground Vehicle (UGV) manipulating an object.
The two units are subject to input saturations and collaborate to move the object to a desired pose characterized by its position and inclination.
The dynamics are derived using Euler-Lagrange method.
A pre-stabilizing control law is proposed where the UGV is tasked to deploy the object to a certain position whereas the UAV adjusts its inclination.
In particular, a proportional-derivative control law is proposed for the UGV, and a cascade control approach is used for the UAV, where the inner loop controls the attitude of the UAV and the outer loop stabilizes the inclination of the object.
Then, we prove the stability of the points of equilibrium using small gain arguments.
To ensure constraints satisfaction at all times, a reference governor unit is added to the pre-stabilizing control scheme.
Finally, numerical results combined with experimental results are provided to validate the effectiveness of the proposed control scheme in practice.

	\end{abstract}

\end{frontmatter}

\section{Introduction}

Unmanned Aerial Vehicles (UAVs) have been so far used for remote sensing to perform \emph{e.g.,} aerial photography \cite{szantoi2017mapping}, monitoring \cite{san2018uav}, and agriculture \cite{angel2017uav}.
In very recent years, the study of their interactions with the environment has attracted the interest of researchers, giving rise to the new field of \emph{Aerial Robotics} \cite{kondak2015unmanned,nguyen2018novel}.
Research on this topic aims at extending the use of UAVs to more complex missions where UAVs physically interact with the environment and with other robots.
Most works on aerial manipulation deal with the transportation of objects through single and multiple UAVs, including works on grasping \cite{ramon2017detection},  and cooperative transportation \cite{loianno2018cooperative,papasideris2017practical}. 
As aerial manipulation became more popular and more sophisticated, it has also triggered other research initiatives such as making UAVs capable to collaborate with Unmanned Ground Vehicles (UGVs).
Early works on the physical interaction between UAVs and UGVs include the pulling of a cart through one or two quadrotors \cite{srikanth2011controlled}, the cooperative pose stabilization of a UAV through a team of ground robots \cite{naldi2012cooperative} and the modeling and control \cite{marco2017,Tognon2017} of tethered UAVs.
To the best of the authors' knowledge, the study of the manipulation of objects using a team of aerial and ground vehicles is still in its early phase of development. 
Very few studies on the subject exist in the literature. 
For example, our preliminary paper \cite{nguyen2016control} proposes a first control law for a UAV and a UGV manipulating an object subject to actuator saturations. In \cite{staub2017towards}, the authors propose a controller for tracking.
This paper makes use of model inversion techniques, which can be problematic in the presence of \emph{e.g.,} model uncertainties and disturbances.
In this paper, we propose to control the cooperative system using a control law (based on proportional-derivative law) which is inherently robust as a highly accurate model is not required to design the controller.
This paper substantially extends the preliminary results proposed by the same authors in \cite{nguyen2016control}.
The results of \cite{nguyen2016control} were derived under the assumption that the sum of the masses of the UAV and of the object were negligible with respect to the UGV mass, resulting in a non Euler-Lagrange system.
In this paper, we discard this limitative assumption, introducing new coupling dynamics and resulting in a more complete and mechanically correct dynamics (directly derived from Euler-Lagrange).
In this context, the stability proofs of \cite{nguyen2016control} become inapplicable and a new input-to-state (ISS) Lyapunov is introduced to prove stability.
The stability results of this paper are new and represent the main contribution of this paper.
Furthermore, this paper provides more extensive simulations combined with new experimental results to validate the model and the effectiveness of the proposed control scheme.

This paper is organized as follows.
First, the complete dynamics of the system are derived using Euler-Lagrange methods. Then, the attainable configurations of equilibrium are computed considering the saturations of the actuators.
Afterwards, a control scheme is proposed where the stability of the points of equilibrium of the system is proved using small gain arguments and strict Lyapunov functions.
To ensure constraints satisfaction at all times, the control scheme is augmented with the Reference Governor (RG).
Finally, numerical and experimental results are compared and discussed.

\section{Problem Statement}
\begin{figure}[!t]
	\centering
	\includegraphics[width=0.7\columnwidth]{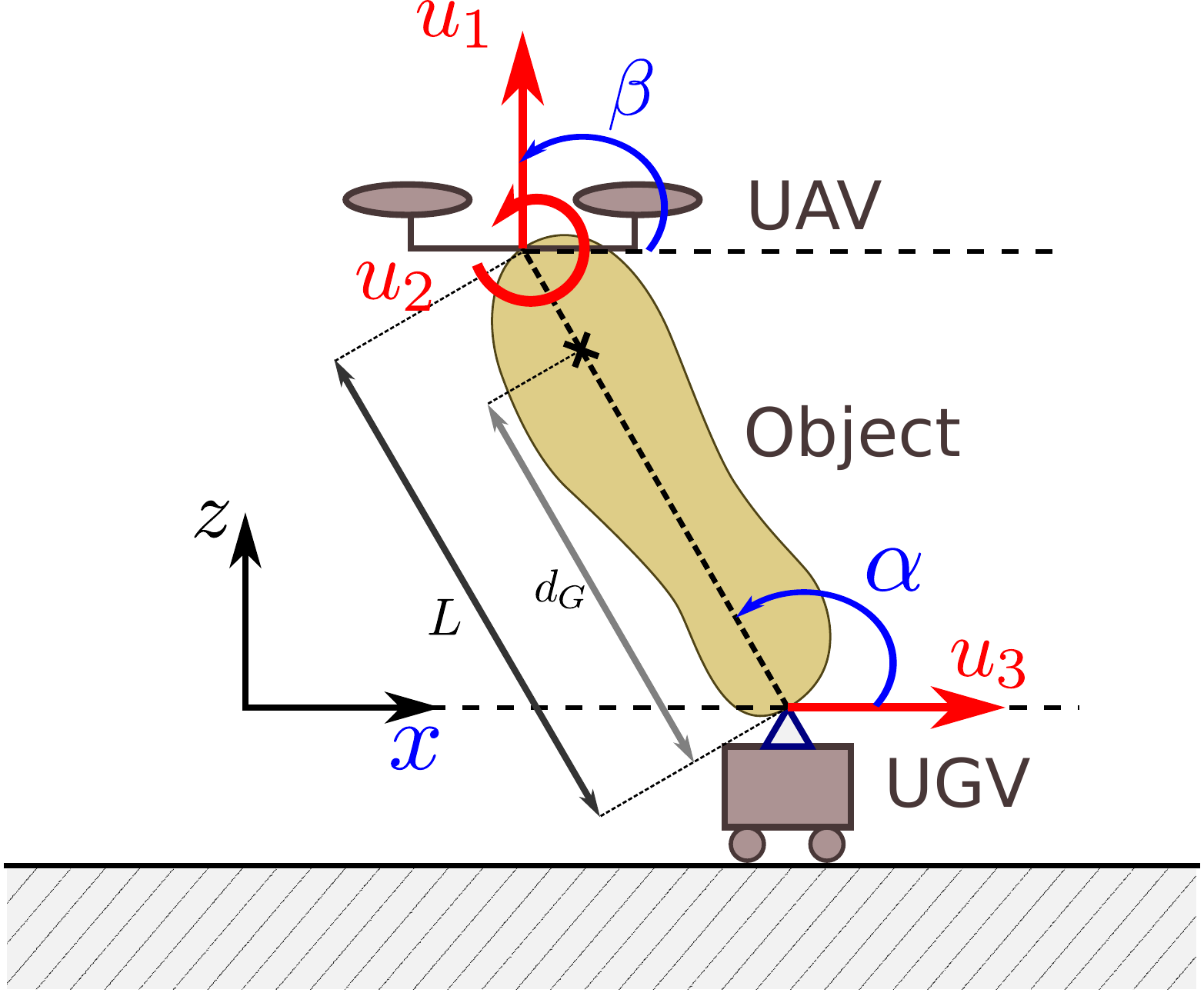}
	\caption{Model of a UAV and a UGV manipulating an object.}\label{model}
\end{figure}

Consider the planar model of a quadrotor UAV\footnote{In this paper, we consider a quadrotor UAV with a particular fixed yaw angle so that the quadrotor can be considered as a birotor in 2D. Note that the thrust and the torque of the birotor in 2D can be mapped to the quadrotor by suitably distributing the forces to the four propellers.} and a UGV manipulating a rigid body (object) as depicted in Fig. \ref{model}. We assume that the center of mass of the UAV coincides with the joint position where it is attached to. 
The UAV has mass $m_a\in\mathbb{R}_{>0}$ and moment of inertia $\mathcal{I}_a\in\mathbb{R}_{>0}$. 
The UGV has mass $m_s\in\mathbb{R}_{>0}$ and the object has mass $m_b\in\mathbb{R}_{>0}$, moment of inertia $\mathcal{I}_b\in\mathbb{R}_{>0}$, and length $L\in\mathbb{R}_{>0}$. 
The center of mass of the object is positioned at distance $d_G\in\mathbb{R}_{>0}$ from the UGV. 
Let the position of the object (attached to the UGV) $x\in\mathbb{R}$, the inclination of the object $\alpha\in[0,\pi]$, and the attitude of the UAV $\beta\in[-\pi,\pi)$ be the generalized coordinates of the system. 
All the angles are defined counter-clockwise with respect to the horizon, and each body is subject to the gravity acceleration $g\in\mathbb{R}_{>0}$. 
The UAV propellers generate the total thrust $u_1\in\mathbb{R}_{\geq 0}$ and the resultant torque $u_2\in\mathbb{R}$. The UGV motors produce the force $u_3\in\mathbb{R}$. 
The signs of the inputs $u_1$, $u_2$, and $u_3$ are defined positive with respect to the oriented vectors depicted in Fig. \ref{model}. The actuators are saturated as
\begin{equation}\label{constraints}
	0\leq u_1\leq T_{max},\quad \lVert u_2 \lVert \leq \tau_{max},\quad
	\lVert u_3 \lVert \leq F_{max},
\end{equation}
where $T_{max}\in\mathbb{R}_{>0}$, $\tau_{max}\in\mathbb{R}_{>0}$, and $F_{max}>T_{max}$.

The equations of motion of the system are derived from Euler-Lagrange. 
Assuming friction forces negligible,
we can derive the equations of motion as
\begin{subequations}\label{systemS}
	\begin{align}
		M_{t}\ddot{x}-M_a L(\ddot{\alpha}\sin\alpha+\dot{\alpha}^2\cos\alpha) =& u_{ff}\label{eq:systemS1}\\
		-M_a L(\ddot{x}\sin\alpha-g\cos\alpha) + \mathcal{I}_0L\ddot{\alpha} =& u_n\sin\theta\label{eq:systemS2}\\
		\mathcal{I}_a\ddot{\beta} =& u_2.\label{eq:systemS3}
	\end{align}
\end{subequations}
where $\theta\triangleq\beta-\alpha$ is the relative angle between the object and the UAV, $M_{t}\triangleq m_s+m_b+m_a$ is the total mass of the system, $M_a \triangleq\frac{m_bd_G}{L}+m_a$ represents the apparent mass of the UAV and the object, and $\mathcal{I}_0\triangleq\frac{m_bd_G^2+\mathcal{I}_b}{L}+m_aL$ is the moment of inertia of the system divided by $L$. Furthermore, we define $u_{ff}\triangleq u_3+u_1\cos\beta$ as the feed-forward control input where $u_1\cos\beta$ is feed-forwarded to $u_3$, and $u_{n}\triangleq u_1L$ as the normalized control input.

It is worth noting that system \eqref{eq:systemS1}-\eqref{eq:systemS2} can be represented as an open-chain robotic manipulator since it can be rewritten as 
	\begin{equation}
		M(q)\ddot{q}+C(q,\dot{q})\dot{q}+g(q) =\tau\label{roboticsystem1},
	\end{equation}
	where $q\triangleq\left[\begin{matrix}
			x\\
			\alpha
	\end{matrix}\right]$, $
	M(q)\triangleq\left[\begin{matrix}
			M_{t} & -M_a L\sin\alpha\\
			-M_a L\sin\alpha & \mathcal{I}_0L
	\end{matrix}\right]$, $
	C(q,\dot{q})\triangleq\left[\begin{matrix}
			0 &-M_a L\dot{\alpha}\cos{\alpha}\\
			0 & 0
			\end{matrix}\right]$, $g(q)\triangleq\left[\begin{matrix}
			0 \\
			M_a Lg\cos\alpha
	\end{matrix}\right]
	$, and $	\tau\triangleq\left[\begin{matrix}
			u_{ff}\\
			u_n\sin\theta
\end{matrix}\right]$ are the coordinates, the inertia, the Coriolis, the gravity, and the external force matrices, respectively.
As detailed in Appendix \ref{app:basicproperties}, system \eqref{roboticsystem1} enjoys the basic properties of open-chain manipulators, \emph{i.e.,} the inertia matrix $M(q)$ is positive definite, and $(\dot{M}(q)-2C(q,\dot{q}))$ is skew-symmetric.

The objective of this paper is to stabilize the pose (\emph{i.e.,} position and orientation) of the object to the desired position $x_d$ and the desired angle $\alpha_d$ by means of the cooperation of the UAV and the UGV. 
Prior to designing the controllers of each unit, we will first analyze the attainable configurations of equilibrium in the presence of input saturations.

\section{Attainable Configurations of Equilibrium}\label{sec:configequilibrium}
In this section, the attainable configurations of equilibrium $[\bar{x} \; \bar{\alpha} \; \bar{\beta}]^T$ and the associated steady-state input vector $\bar{u}=[\bar{u}_1 \; \bar{u}_2 \; \bar{u}_3]^T$ are computed taking into account constraints \eqref{constraints}. 
Setting all the time derivatives of (\ref{systemS}) to zero, it follows that the configurations of equilibrium must satisfy the system of equations
\begin{subequations}\label{equilibrium1}
	\begin{align}
		\bar{u}_3&=-\bar{u}_1\cos\bar{\beta}\label{eq:equilibrium1first},\\
		\bar{u}_1\sin(\bar{\beta}-\bar{\alpha})&=M_a g\cos\bar{\alpha}\label{eq:equilibrium1second},\\
		\bar{u}_2&=0.\label{eq:equilibrium1third}
	\end{align}
\end{subequations}
Clearly, Eq. \eqref{eq:equilibrium1third} gives $\bar{u}_2=0$ as the only attainable input associated to an equilibrium. 
Moreover, note that any position $\bar{x}\in\mathbb{R}$ is an attainable point of equilibrium since $\bar{x}$ does not appear in (\ref{equilibrium1}).
Regarding \eqref{eq:equilibrium1first}, since we assumed that $F_{max}>T_{max}$, the force $\bar{u}_3$ at equilibrium always exists for any $\bar{u}_1\leq T_{max}$ and $\bar{\beta}\in [-\pi,\pi)$.
For what concerns Eq. \eqref{eq:equilibrium1second}, it is possible to compute the maximum value of $\bar{u}_1\sin(\bar{\beta}-\bar{\alpha})$ using the fact that $\bar{u}_1\leq T_{max}$. In particular, the maximum value is reached when $\bar{u}_1=T_{max}$ and $\bar{\beta}=\bar{\alpha}\pm\pi/2$.
Accordingly, there are two possible cases:
\begin{itemize}
	\item If $T_{max}\geq M_a g$, then any $\bar{\alpha}\in[0,\pi]$ is an attainable angle of equilibrium;
	\item If $T_{max}<M_a g$, then the attainable angles of equilibrium are restricted to the interval $\bar{\alpha} \in [\alpha_{min}, \alpha_{max}]$, where the boundaries are $\alpha_{min}=\arccos\left({\frac{T_{max}}{M_a g}}\right)$ and $\alpha_{max}=\arccos\left(\frac{-T_{max}}{M_a g}\right)$.
\end{itemize}

Finally note that, for a given steady-state angle $\bar{\alpha}$, the attainable equilibria for the attitude $\bar{\beta}$ are restricted to the interval $\bar{\beta} \in [\beta_{min},\beta_{max}]$. The boundaries of this interval can be computed solving \eqref{eq:equilibrium1second} by substituting $\bar{u}_1=T_{max}$. Doing so, we obtain 
$\beta_{min}= \arcsin{\left(\frac{\pm M_a g\cos{\bar{\alpha}}}{T_{max}}\right)}+\bar{\alpha}$ and $\beta_{max}= \pi-\arcsin{\left(\frac{\pm M_a g\cos{\bar{\alpha}}}{T_{max}}\right)}+\bar{\alpha}$, where the positive sign is taken if $\bar{\alpha}\in [\alpha_{min},\pi/2)$, and the negative one if $\bar{\alpha}\in [\pi/2,\alpha_{max}]$.

\section{Control Scheme}

\begin{figure}[!t]\centering{
		\includegraphics[width=0.75\columnwidth]{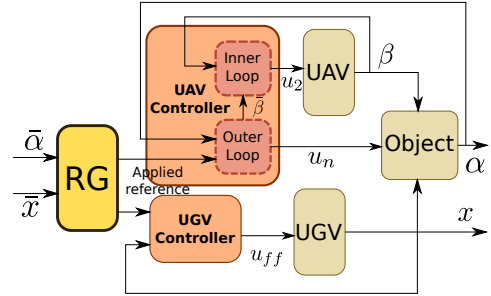}
	\caption{Proposed control scheme.} \label{controlarchitecture}}
\end{figure}
The proposed control scheme consists of two separate control units that are controlling the UAV and the UGV (see Fig. \ref{controlarchitecture}).
The UGV controller generates the control input $u_{ff}$ so as to make the object position asymptotically tend to $x_d$.
The UAV controller uses a cascade control approach, where the inner loop is tasked with the control of the UAV attitude $\beta$, whereas the outer loop is tasked to control the inclination $\alpha$ of the transported object.
For constraints satisfaction, an RG unit is added to the scheme. Whenever necessary, the RG modifies the desired references $\alpha_d$ and $x_d$ to $\alpha_a$ and $x_a$, respectively, to ensure that constraints are satisfied at all times.
In the following subsections, the control laws of the UGV and the UAV are detailed.

\subsection{UGV Control Law}

The objective of the UGV control law is to steer the object to the desired position $x_d$. 
For this purpose, the following proportional-derivative (PD) control law is proposed:
\begin{equation}\label{u3}
	u_{ff}=k_{p,x}\tilde{x}-k_{v,x}\dot{x},
\end{equation}
where $\tilde{x}\triangleq x_d-x$ is the position error, and $k_{p,x},k_{v,x}\in\mathbb{R}_{>0}$ are the control gains to be tuned.

\subsection{UAV Control Law}
The UAV control law uses a cascade control approach, where the inner loop controls the UAV attitude and the outer loop controls the inclination of the object.

\subsubsection{Inner Loop}

To control the UAV attitude, a PD control law is chosen:
\begin{equation}\label{u2control}
	u_2=-k_{p,\beta}\tilde{\beta}-k_{v,\beta}\dot{\beta},
\end{equation}
where $k_{p,\beta},k_{v,\beta}\in\mathbb{R}_{>0}$ are control parameters to be tuned, $\tilde{\beta}\triangleq\beta_d-\beta$ is the attitude error and $\beta_d$ the desired UAV attitude. 

\subsubsection{Outer Loop}

We define $\theta_d\triangleq \beta_d-\alpha$ as the desired relative attitude of the UAV, and $f_t$ the tangential force produced by the UAV on the object as
\begin{equation}\label{ft}
	f_t\triangleq u_n\sin{\theta_d}.
\end{equation}
For the moment, assume that we can use $f_t$ as a new control input to stabilize the inclination of the object $\alpha$; the proposed control law is a PD with gravity compensation
\begin{equation}\label{ftexplicit}
	f_t=k_{p,\alpha}\tilde{\alpha}-k_{v,\alpha}\dot{\alpha}+M_a Lg\cos{\alpha},
\end{equation}
where $\tilde{\alpha}\triangleq \alpha_d-\alpha$ is the object inclination error, and $k_{p,\alpha},k_{v,\alpha}\in\mathbb{R}_{>0}$ are the parameters to be tuned so that $k_{p,x}>k_{p,\alpha}$ and $k_{v,x}>k_{v,\alpha}$.
It remains to construct $u_n$ and $\theta_d$ that produce the desired tangential force \eqref{ftexplicit}. 
In line of principle, Eq. (\ref{ft}) admits an infinite number of solutions for $u_n$ and $\theta_d$. 
However, in this paper, we propose the following continuous mapping\footnote{We define the saturation function as $\sigma_\lambda(x)\triangleq\text{\normalfont{sign}}(x)\min{(|x|,\lambda)}, \quad \lambda>0$}
\begin{equation}\label{maptheta}
	\theta_d=\sigma_{\pi/2}(\gamma\arctan{(\epsilon f_t)}),
\end{equation}
where $\gamma,\epsilon\in\mathbb{R}_{>0}$ are parameters to be chosen such that thrust constraints are satisfied at steady-state. 
The main advantage of (\ref{maptheta}) is that this mapping always guarantees the positiveness of $u_n$. Indeed, both functions $f_t$ and $\sin(\sigma_{\pi/2}(\gamma\arctan{(\epsilon f_t)}))$ are odd and monotonically increasing with respect to the variable $f_t$. Hence, rewriting (\ref{ft}) as
\begin{equation}\label{mapu1}
	u_n=\dfrac{f_t}{\sin{\theta_d}},
\end{equation}
the resulting $u_n$ is always positive since the quotient of two odd and monotonically increasing functions is always positive. 
Another relevant property of (\ref{maptheta}) is that $u_n$ does not present any singularities since for any $f_t$ not equal to zero, \eqref{maptheta} is always determined, and $\lim_{f_t\to0}u_n=1/(\gamma\epsilon)$. Finally, the mapping (\ref{maptheta}) presents the interesting advantage that, if the parameter $\gamma$ is properly chosen, we can prove that it is possible to freely choose the parameter $\epsilon$ (becoming then a tuning parameter) which ensures constraints satisfaction at any attainable configuration of equilibrium. In particular, if we choose $\gamma$ so as to satisfy saturations (\ref{constraints}) when $f_t=T_{max}L$, it follows from (\ref{mapu1}) that $\theta_d=\pi/2$. As a result, following from (\ref{maptheta}), we obtain
\begin{equation}\label{gamma}
	\gamma = \dfrac{\pi}{2\arctan(\epsilon T_{max}L)}.
\end{equation}
As clarified in the following Lemma, with $\gamma$ fixed as in (\ref{gamma}), steady-state constraints are always ensured for any $\epsilon\in\mathbb{R}_{>0}$ and, as a consequence, $\epsilon$ can be freely chosen as a tuning parameter.

\begin{lem}\label{epsilontuning}
	For any $\epsilon\in\mathbb{R}_{>0}$, the mapping (\ref{maptheta}) with $\gamma$ satisfying (\ref{gamma}) ensures $|\bar{u}_1|\leq T_{max}$ at equilibrium.
\end{lem}
\begin{pf}
	Consider first the particular case where $f_t\in[0,T_{max}L]$. 
	In view of (\ref{ftexplicit}), the control input $f_t$ at equilibrium must satisfy $f_t=M_a Lg\cos\bar{\alpha}$. 
	Defining the minimum relative UAV attitude $\bar{\theta}_{min}\triangleq\beta_{min}-\bar{\alpha}$, and using $\beta_{min}$ found in Section \ref{sec:configequilibrium}, we obtain $\bar{\theta}_{min}=\arcsin(f_t/(T_{max}L))$.
	Then, following from \eqref{eq:equilibrium1second} and defining $\bar{\theta}=\bar{\beta}-\bar{\alpha}$, to ensure $|\bar{u}_1|\leq T_{max}$ for all points of equilibrium, $\bar{\theta}$ must satisfy $\bar{\theta} \geq \bar{\theta}_{min}$.
	As a consequence, the inequality $\bar{\theta}\geq\bar{\theta}_{min}$ must be satisfied for any $f_t\in[0,T_{max}L]$. 
	Choosing $\gamma$ as in (\ref{gamma}), the inequality $\gamma\arctan{(\epsilon f_t)}\geq\arcsin(f_t/(T_{max}L))$
	holds true for any $\epsilon\in\mathbb{R}_{>0}$ since, if restricted to $f_t\in[0,T_{max}L]$, $\gamma\arctan{(\epsilon f_t)}$ is convex and $\arcsin(f_t/(T_{max}L))$ is concave.
	The same arguments hold true for any $f_t\in[-T_{max}L,0]$, where $\bar{\theta}\leq\bar{\theta}_{max}$ with $\bar{\theta}_{max}\triangleq\beta_{max}-\bar{\alpha}$, concluding the proof.
\end{pf}

\section{Stability Properties}\label{sec:stabilityproperties}

This section is dedicated to prove the asymptotic stability of \eqref{eq:systemS3},(\ref{roboticsystem1}) using the control law (\ref{u3}), (\ref{u2control}), \eqref{ftexplicit}, (\ref{maptheta}), and (\ref{mapu1}) for $\alpha_d\in(\alpha_{min},\alpha_{max})$.
In order to do so, we first prove that \eqref{eq:systemS3} controlled by (\ref{u2control}) is ISS with respect to $\dot{\beta}_d$.
Furthermore, we prove that the asymptotic gain $\gamma_{in}$ of the inner loop can be made arbitrarily small by acting on $k_{p,\beta}$ and $k_{v,\beta}$.
Afterwards, (\ref{roboticsystem1}) controlled by (\ref{u3}), \eqref{ftexplicit}, (\ref{maptheta}), and (\ref{mapu1}) is proved to be ISS with restriction with respect to $\tilde{\beta}$.
This enables to prove that the asymptotic gain $\gamma_{out}$ of the outer loop exists and is finite.
As a consequence, since the overall system is interconnected, it is possible to prove that the points of equilibrium are asymptotically stable using small gain arguments.

\vspace{-0.2cm}
To prove that the inner loop is ISS with respect to $\dot{\beta}_d$, we reformulate the inner attitude dynamics of \eqref{eq:systemS3} controlled by (\ref{u2control}) as 
\begin{subequations}\label{attitudedynamics}
	\begin{align}
		\dot{\tilde{\beta}}&=\dot{\beta}-\dot{\beta}_d\\
		\mathcal{I}_a\ddot{\beta}&=-k_{p,\beta}\tilde{\beta}-k_{v,\beta}\dot{\beta}.
	\end{align}
\end{subequations}
System (\ref{attitudedynamics}) represents the dynamics of the inner loop, where the states $[\tilde{\beta} \; \dot{\beta}]^T$ are affected by the exogenous input $\dot{\beta}_d$.
The following property can be proved.
\begin{lem}\label{UAVAttitudeProposition}
	The inner-loop (\ref{attitudedynamics}) is ISS with respect to $\dot{\beta}_d$ for any $k_{p,\beta}>0$ and $k_{v,\beta}>0$. The asymptotic gain $\gamma_{in}$ between the disturbance $\dot{\beta}_d$ and the output $\tilde{\beta}$ is finite and can be made arbitrarily small for sufficiently large $k_{p,\beta}>0$ and $k_{v,\beta}>0.$
\end{lem}
\begin{pf}
	Please refer to \cite[Proposition 15]{nicotra2014taut}.
\end{pf}
The next step is to prove that (\ref{roboticsystem1}) controlled by (\ref{u3}), \eqref{ftexplicit}, (\ref{maptheta}) and (\ref{mapu1}) is ISS with restriction with respect to $\tilde{\beta}$ and that there is a finite gain $\gamma_{out}$ between $\tilde{\beta}$ and $\dot{\beta}_d.$ To do so, let us first rewrite the control action  (\ref{u3}), \eqref{ftexplicit}, (\ref{maptheta}) and (\ref{mapu1}) so as to isolate  the effect of the attitude error $\tilde{\beta}=\tilde{\theta}$. Since $\theta=\theta_d+\tilde{\theta}$, we can use (\ref{ft}) and (\ref{ftexplicit}) to rewrite the right-hand side of \eqref{eq:systemS2} as
\begin{align}\label{passage6}
	u_n\sin\theta= (k_{p,\alpha}\tilde{\alpha}-k_{v,\alpha}\dot{\alpha})\cos\tilde{\theta} + g_2(\alpha)+\delta_{\tilde{\theta},2},
\end{align}
where $g_2(\alpha)\triangleq M_a Lg\cos\alpha $ is the second component of $g(q)$, and $\delta_{\tilde{\theta},2}\triangleq f_t\dfrac{\cos\theta_d}{\sin\theta_d}\sin\tilde{\theta}-2g_2(\alpha)\sin^2(\tilde{\theta}/2)$.
At this point, using (\ref{u3}) and (\ref{passage6}), the external force vector $\tau$ can be rewritten as
\begin{align}\label{passage7}
	\tau = K_p(\tilde{\theta})\tilde{q}-K_v(\tilde{\theta})\dot{q}+g(q)+ \delta_{\tilde{\theta}},
\end{align}
where $\tilde{q}\triangleq\left[\begin{matrix}\tilde{x} \\ \tilde{\alpha}\end{matrix}\right]$ , $K_p(\tilde{\theta})\triangleq\left[\begin{matrix}k_{p,x} & 0 \\ 0 & k_{p,\alpha}\cos\tilde{\theta}\end{matrix}\right]$ , $K_v(\tilde{\theta})\triangleq\left[\begin{matrix}
	k_{v,x} & 0\\
	0 & k_{v,\alpha}\cos\tilde{\theta}
\end{matrix}\right]$, $\delta_{\tilde{\theta}}\triangleq\left[\begin{matrix} 0 \\ \delta_{\tilde{\theta},2}\end{matrix}\right]$ are the state error, the proportional gain, the derivative gain, and the exogenous input matrices (depending on $\tilde{\theta}$) affecting the states $[\tilde{q}^T \; \dot{q}^T]^T$ of (\ref{roboticsystem1}), respectively.
Interestingly enough, it is possible to prove that $\|\delta_{\tilde{\theta}}\|$ can be upper-bounded by a saturated linear function of $\tilde{\theta}.$ 
\begin{lem}\label{u1coslemma}
	The norm of $\delta_{\tilde{\theta}}$ satisfies $\lVert \delta_{\tilde{\theta}} \lVert \leq \max \{(2/ \pi+2M_a Lg) |\tilde{\theta}|,2/ \pi+2M_a Lg\}$ 
	for any $\epsilon\in\mathbb{R}_{>0}$, $f_t\in\mathbb{R},$ $\alpha \in [0,\pi]$, and $\tilde{\theta}\in[-\pi,\pi)$.
\end{lem}
\begin{pf}
	Using triangular inequality, it is possible to prove that $|f_t\cot\bar{\theta}\sin\tilde{\theta}|\leq 2/\pi|\tilde{\theta}|$ and that $2M_aLg|\cos\alpha|\sin^2(\tilde{\theta}/2)\leq 2M_aLg|\tilde{\theta}|$.
	As a result, the property stated by Lemma \ref{u1coslemma} holds true. 
	For more details, please refer to Appendix \ref{u1coslemmaproof}.
\end{pf}
The following Lemma proves that (\ref{roboticsystem1}) controlled by (\ref{u3}), \eqref{ftexplicit}, (\ref{maptheta}), and (\ref{mapu1}) is ISS with restriction with respect to the attitude error $\tilde{\beta}$ and that there exists a finite asymptotic gain $\gamma_{out}$ between $\tilde{\beta}$ and $\dot{\beta}_d.$
\begin{lem}\label{outerloopISS}
	Given the desired position $x_d \in \mathbb{R}$, the desired inclination $\alpha_d \in (\alpha_{min}, \alpha_{max})$, and the resulting steady-state attitude $\beta_d$ resulting from (\ref{maptheta}), system (\ref{roboticsystem1}) controlled by (\ref{passage7}) is ISS with restriction $|\tilde{\theta}|<\tilde{\theta}_{max}$ (or equivalently $|\tilde{\beta}| < \tilde{\theta}_{max}$), and $\lVert \tilde{q} \lVert < \tilde{q}_{max}$ with respect to $\tilde{\beta}.$ Furthermore the asymptotic gain $\gamma_{out}$ between the disturbance $\tilde{\beta}$ and the output $\dot{\beta}_d$ exists and is finite\footnote{For a complete characterization of $\tilde{q}_{max}$, $\tilde{\theta}_{max}.$ and $\gamma_{out}$, please refer to \eqref{qmaxfinal}, \eqref{thetafinal}, and \eqref{gammaout}, respectively, in Appendix \ref{outerloopISSproof}.}.
\end{lem}
\begin{pf}
		The proof uses the Lyapunov function \cite{santibanez1997strict}
		\begin{equation}\label{lyapunov}
			\begin{aligned}
				V=\dfrac{1}{2}\dot{q}^TM(q)\dot{q}+U_T(q_d,\tilde{q},0)-\gamma_p f(\tilde{q})^TM(q)\dot{q},
			\end{aligned}
		\end{equation}
		where $U_T(\tilde{q},\tilde{\theta})\triangleq 1/2 \tilde{q}^TK_p(\tilde{\theta})\tilde{q}$,
		$f(\tilde{q})\triangleq\tilde{q}/(1+\lVert\tilde{q}\lVert)$,
		$\gamma_p>0$ is a positive scalar satisfying $\gamma_p <\min\left\{\sqrt{\frac{2b}{\lambda_M\{M\}}},\frac{2b'}{\lambda_M\{K_v(\tilde{\theta})\}},\frac{\lambda_m\{K_v(\tilde{\theta})\}}{2(k_c+2\lambda_M\{M\})}\right\}$,
		$\lambda_M\{M\}$ is the largest eigenvalue of $M$, $\lambda_M\{K_v(\tilde{\theta})\} \allowbreak = k_{v,x}$ is the largest eigenvalue of $K_v(\tilde{\theta})$, and $k_c>0$ is a positive scalar satisfying $\lVert C(q,\dot{q})\dot{q}\lVert\leq k_c\lVert\dot{q}\lVert^2$.
		It can be shown that the time derivative of $V$ can be bounded by
		\begin{equation*}\label{proof10}
			\begin{aligned}
				\dot{V}&\leq-\dfrac{1}{2}[\dot{q}-\gamma_p f(\tilde{q})]^TK_v(\tilde{\theta})[\dot{q}-\gamma_p f(\tilde{q})]\\
				       &-[\dfrac{1}{2}\lambda_m\{K_v(\tilde{\theta})\}-2\gamma_p\lambda_M\{M\}-\gamma_p k_c] \lVert \dot{q} \lVert ^2\\
				       &+\dot{q}^T[K_p(\tilde{\theta})-K_p(0)]\tilde{q}
				       -\dfrac{\gamma_p}{1+ \lVert \tilde{q} \lVert }[\tilde{q}^TK_p(\tilde{\theta})\tilde{q}\\
				       &-\dfrac{\gamma_p \tilde{q}^T K_v(\tilde{\theta})\tilde{q}}{2(1+ \lVert \tilde{q} \lVert )}]
				       +[\dot{q}-\gamma_p f(\tilde{q})]^T\delta_{\tilde{\theta}}.
			\end{aligned}
		\end{equation*}
		To prove ISS with restriction, it is enough to use the fact that $\dot{V}<0$ whenever $\lVert \tilde{q},\dot{q} \lVert$ remains outside a ball of radius $\rho|\tilde{\theta}|$ with restriction $\lVert \tilde{q} \lVert < \tilde{q}_{max}$ and $| \tilde{\theta} | < \tilde{\theta}_{max}$. 
		The details of the proof can be found in Appendix \ref{outerloopISSproof}.
\end{pf}
\vspace{-0.2cm}
Combining Lemmas \ref{UAVAttitudeProposition} and \ref{outerloopISS}, it is possible to prove the asymptotic stability of the points of equilibrium.
\begin{thm}\label{smallgainproposition}
	Consider system \eqref{eq:systemS3},(\ref{roboticsystem1}) controlled by (\ref{u3}), (\ref{u2control}), (\ref{ftexplicit}), and (\ref{maptheta}).
	Given the desired position $x_d \in \mathbb{R},$ the desired inclination $\alpha_d \in (\alpha_{min}, \alpha_{max}),$ and the resulting steady-state attitude $\beta_d$, the point of equilibrium $[\bar{x} \; \bar{\alpha} \; \bar{\beta}]^T=[x_d \; \alpha_d \; \beta_d]^T$  is asymptotically stable for suitably large $k_{p,\beta}$ and $k_{v,\beta}$. 
\end{thm}
\begin{pf}
	From Lemmas \ref{UAVAttitudeProposition} and \ref{outerloopISS}, $\gamma_{in}$ and $\gamma_{out}$ are proved to be finite under the assumption $|\tilde{\theta}|<\tilde{\theta}_{max}$ and $\lVert \tilde{q} \lVert < \tilde{q}_{max}$. 
	In this case, we can achieve $\gamma_{in}\gamma_{out}<1$ since $\gamma_{in}$ can be made arbitrarily small for sufficiently large $k_{p,\beta}$ and $k_{v,\beta}$. 
	Therefore, the Small Gain Theorem can be applied and, for a suitable set of initial conditions around the point of equilibrium that satisfies $\lVert \tilde{\theta} \lVert_\infty < \tilde{\theta}_{max}$ and $\lVert \tilde{q} \lVert_\infty < \tilde{q}_{max}$, the closed loop system is asymptotically stable.
\end{pf}
Interestingly enough, it is possible to improve this control law by substituting (\ref{mapu1}) with\footnote{In this paper, we define the \emph{positive} saturation function as $\sigma_{0,\lambda}(x)\triangleq \max\{\sigma_{\lambda}(x),0\}$.}
\begin{equation}\label{u1new}
	u_n=\sigma_{0,T_{max}L}\left(\dfrac{f_t}{\sin\theta}\right).
\end{equation}
The main difference between (\ref{mapu1}) and \eqref{u1new} is that, instead of dividing $f_t$ by the desired relative attitude $\theta_d$, the new control law (\ref{u1new}) divides $f_t$ directly by the actual relative attitude $\theta$.
The following proposition proves that the new control law (\ref{u1new}) preserves the previous stability results.
\begin{prop}\label{propositionu1new}
	Consider \eqref{eq:systemS3},(\ref{roboticsystem1}) controlled by (\ref{u3}), (\ref{u2control}), (\ref{ftexplicit}), (\ref{maptheta}) and (\ref{u1new}). 
	For any $\bar{x} \in \mathbb{R}$, for any $\bar{\alpha} \in (\alpha_{min}, \alpha_{max})$, and for the resulting steady-state $\beta_d$, the point of equilibrium $[\bar{x} \; \bar{\alpha} \; \bar{\beta}]^T=[x_d \; \alpha_d \; \beta_d]^T$  is asymptotically stable. Furthermore, the control law (\ref{u1new}) is equivalent to the control law (\ref{mapu1}) with a feed-forward action.
\end{prop}
\begin{pf}
	The control law \eqref{u1new} is equivalent to a feedforward block that possibly reduces the effect of the attitude error on the outer loop.
	In particular, the feedforward block modifies the actual attitude error $\tilde{\theta}$ to $\tilde{\theta}_f$.
	It can be shown that $|\tilde{\theta}_f|\leq |\tilde{\theta}|$, which makes the gain between $\tilde{\theta}$ and $\tilde{\theta}_f$ smaller than one.
	As a result, all the stability results of Theorem \ref{smallgainproposition} apply.
	For more details, please refer to Appendix \ref{proofu1new}.
\end{pf}
\vspace{-0.2cm}
Note that the stability results presented in this section are local (\emph{i.e.,} valid for any initial condition sufficiently close to the equilibrium point). In the next section, the presented control scheme will be augmented with an RG, which makes the system asymptotically stable for a larger set of initial conditions, \emph{e.g.} for any steady-state admissible initial condition satisfying $\alpha(0)\in(\alpha_{min},\alpha_{max})$, and $\beta(0)\in(\beta_{min},\beta_{max})$ with zero initial velocities.

\section{Constraints Enforcement}

In this section, the control law previously studied is augmented with the RG introduced in \cite{bemporad1998reference} to avoid constraints violation and, consequently, increase the basin of attraction of the points of equilibrium. 
Let the desired position and angle references $[x_d,\alpha_d]$ be given, where $x_d\in\mathbb{R}$ and $\alpha_d\in(\alpha_{min},\alpha_{max})$. 
If needed, the RG substitutes the desired set-point $[x_d,\alpha_d]$ with a sequence of applied way-points $[x_a,\alpha_a]_k$ that ensures that the system trajectories do not violate the constraints. 
This sequence is computed online by assuming that, at time $t=k$, the applied reference $[x_a, \alpha_a]_k$, if maintained constant, would not make the system violate the constraints. The RG computes (at fixed time intervals) the next applied reference
$[x_a,\alpha_a]_{k+1}=(1-c)[x_a,\alpha_a]_k+c[x_d,\alpha_d]$
by maximizing the scalar $c \in [0\,\, 1]$ so that, if this reference were to remain constant, the system trajectories would not violate constraints at any future time instant. 
The optimization of $c$ can be performed using bisection \cite{bemporad1998reference} and online simulations over a sufficiently long prediction horizon. 
The convexity of the set of steady-state admissible equilibria ensures that, if $[x_a,\alpha_a]$ is kept constant, the way-point sequence converges to $[x_d,\alpha_d]$.

\section{Simulations}

\begin{figure}[t]\centering
	\includegraphics[width=.9\columnwidth]{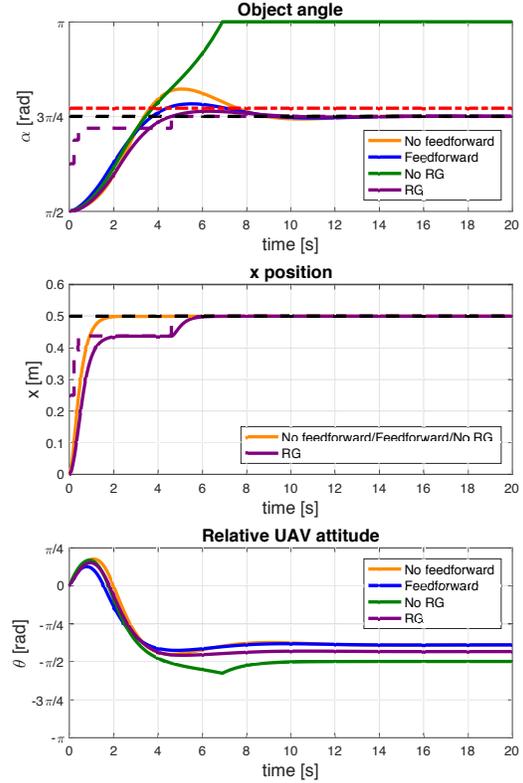}
	\caption{Numerical simulations. The solid lines represent the states of the system, the red dash-dot line represents $\alpha_{max}$, and the dashed lines represent the applied references.}
	\label{firstcontroller}
\end{figure}

Consider a UAV of mass $m_a=100\;\text{[g]}$ and inertia $\mathcal{I}_a=1.014\;\text{[g.m$^2$]}$ cooperating with a UGV of mass $m_s=1\;\text{[kg]}$ to manipulate an object of mass $m_b=30\;\text{[g]}$, length $L=1.25\;\text{m}$, inertia $\mathcal{I}_b=2\;\text{[kg.m$^2$]}$, and whose center of mass is located at $d_G=0.625 \;[\text{m}]$ from the UGV. 
The saturations of the UAV torque and the UGV force are set to $\tau_{max}=0.2\;\text{[Nm]}$ and $F_{max}=20\;\text{[N]}$, respectively. 
Additional viscous frictions are added to the model. The coefficients of viscosity for each general coordinate $x$, $\alpha$, and $\beta$ are $\zeta_x=1.5\;[\text{Ns/m}]$, $\zeta_{\alpha}=0.1\;[\text{Nms}]$, and $\zeta_{\beta}=0.05\;[\text{Nms}]$, respectively.
These parameters have been identified on the experimental testbed presented in the next section.
The system is controlled by (\ref{u3}), (\ref{u2control}), \eqref{ftexplicit}, and (\ref{maptheta}), where the control gains are $k_{p,x}=20$, $k_{v,x}=8.5$, $k_{p,\alpha}=1$, $k_{v,\alpha}=1.5$, $k_{p,\beta}=3$, $k_{v,\beta}=2.85$, and $\epsilon=2.4$. 
Simulations are carried out considering the initial condition $[x(0) \; \allowbreak\dot{x}(0) \; \allowbreak\alpha(0) \; \allowbreak\dot{\alpha}(0) \; \allowbreak\beta(0) \; \allowbreak\dot{\beta}(0)]^T=[0 \; \allowbreak0 \; \allowbreak\pi/2 \; \allowbreak0 \; \allowbreak\pi/2 \; \allowbreak0]^T$ with the desired object pose set to $\alpha_d=3\pi/4$ and $x_d=0.5\;[\text{m}]$. 
Fig. \ref{firstcontroller} provides the numerical comparisons given: 
\begin{itemize}
	\item \textbf{No feedforward:} The maximum thrust of the UAV is set to $T_{max}=5\;[\text{N}]$. The outer loop is controlled by (\ref{mapu1}) and the closed-loop is subject to a direct step variation of the desired reference.
	\item \textbf{Feedforward:} The maximum thrust is set to $T_{max}=5\;[\text{N}]$. The outer loop uses the feedforward \eqref{u1new} instead of \eqref{mapu1}, and the closed-loop is subject to a direct step variation of the desired reference.
	\item \textbf{No RG:} The maximum thrust is reduced to $T_{max}=0.85\;[\text{N}]$. The outer loop uses feedforward \eqref{u1new} and the closed-loop is subject to a direct step variation of the desired reference.
	\item \textbf{RG:} The maximum thrust is set to $T_{max}=0.85\;[\text{N}]$ and the outer loop uses the feedforward \eqref{u1new}. The applied reference is issued by the RG using bisection with sampling time $t_s=0.2\;\text{[s]}$ and a prediction time horizon $t_h=15\;[\text{s}]$.
\end{itemize}

Fig. \ref{firstcontroller} shows that the control law (\ref{mapu1}) stabilizes the system to the desired references but may trigger undesired behaviors such as high overshoots, which can be problematic in the case of less capable UAVs as clarified later on. 
We thus illustrate the advantage of using the feedforward action (\ref{u1new}) instead of the proposed law \eqref{mapu1}.
In particular, we see that using directly the relative angle $\theta$ instead of the desired angle $\theta_d$ to compute $f_t$ can improve the damping performance of the controlled system.
Using the ISE/IAE performance indices \cite{soni2013bf} for the error on $\alpha$, we obtain $\text{ISE/IAE}=1.18/2.47$ without feedforward, and $\text{ISE/IAE}=0.91/1.93$ with feedforward.
\newline
For what regards constraints satisfaction, we compare the response with/without RG to illustrate the utility of implementing the RG in the case of less capable UAVs (\emph{i.e.,} for lower $T_{max}$). Without the use of the RG, we observe in Fig. \ref{firstcontroller} that the system violates the constraints \emph{i.e.,} the object angle $\alpha$ goes beyond $\alpha_{max}$ making the object fall down to $\alpha=\pi$, from which the system cannot recover. 
This is why, the RG is implemented and we observe from Fig. \ref{firstcontroller} that the system trajectories move safely to the desired reference without violating constraints. 

\section{Experimental Results}

\begin{figure}
	\centering
	\includegraphics[width=.7\columnwidth]{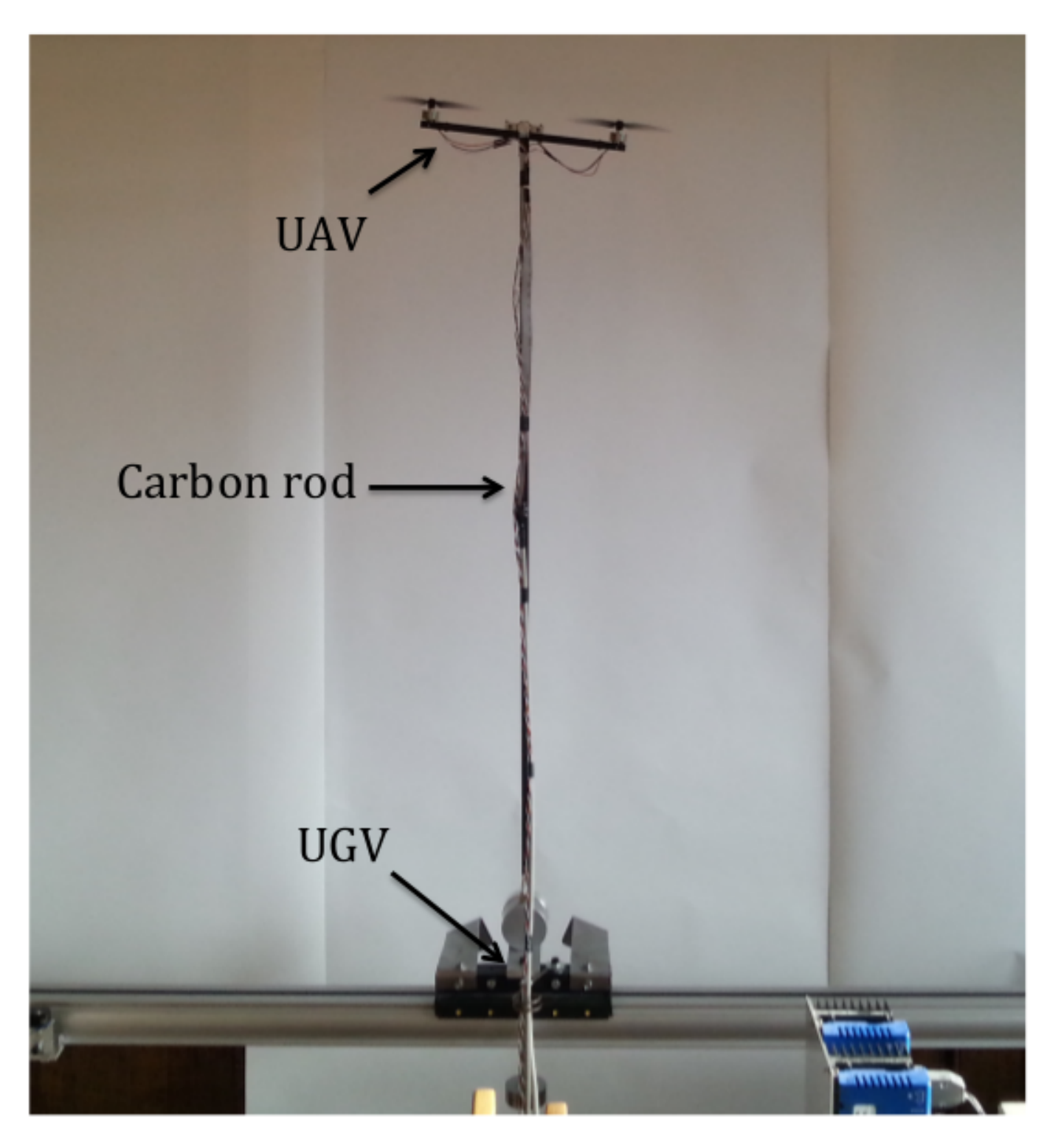}
	\caption{Experimental setup.}\label{fig:expsetup}
\end{figure}

The experimental testbed (see Fig. \ref{fig:expsetup}) consists of a birotor UAV, a carbon rod, and a moving cart.
The encoder \emph{Scancon SCA16} measures the relative attitude of the UAV $\theta$, the \emph{Hengstler Incremental Push Pull Rotary Encoder RI58-0} measures the angle of the rod $\alpha$, and the \emph{Hohner encoder} measures the object position $x$. 
The \emph{Brushless Controller Simon Serie} are electronic speed controllers that control the speed of the UAV propellers, which are brushless DC motors. 
The variable-frequency drive \emph{Junus JSP-090-20} is used to control the \emph{Parvex Axem} motor of the worm drive system. 
The control algorithms are implemented through \emph{dSpace} and \emph{Simulink}. 
The system is controlled using (\ref{u3}), (\ref{u2control}), \eqref{ftexplicit}, (\ref{maptheta}), and \eqref{u1new} using the same parameters as the simulations.
Note that a saturated integral term with gain $k_i=0.001$ has been added to the outer loop to reject the steady-state disturbances induced by some of the neglected aspects, such as the additional gravity term induced by the cables at different positions. 
\begin{figure}[t]\centering
	\includegraphics[width=.9\columnwidth]{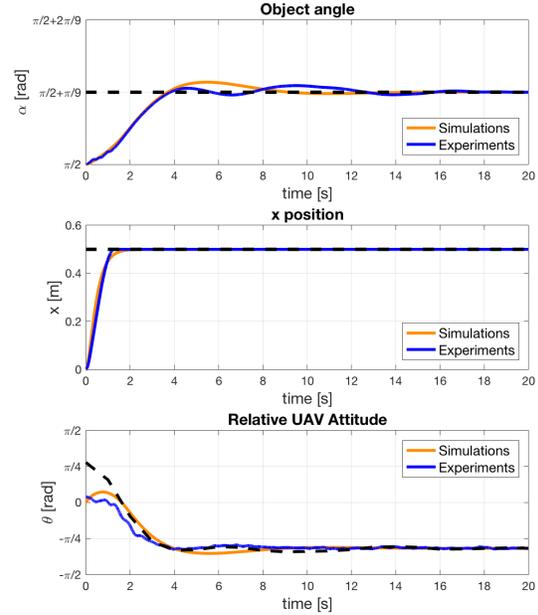}
	\caption{Comparison between numerical results and experimental results. The solid lines represent the states of the system and the dashed lines represent the applied reference.}\label{expplots}
\end{figure}
The 
desired references for the object inclination and position are set to $\alpha_d=\pi/2+\pi/9$ and $x_d=0.5\;[\text{m}]$, respectively. 
Fig. \ref{expplots} shows that the proposed control law stabilizes the system to the desired references. 
Furthermore, we show that the data from the simulated model fits well the real data. 
The videos of the experiments can be found on \url{https://wp.me/p9eDF3-3W}.

\section{Conclusions}

This paper proposes a control scheme to position and orient an object by means of a UAV and a UGV. 
In particular, this paper proposes a control scheme where the UAV is tasked with the control of the object inclination, whereas the UGV is tasked with the control of the object position. 
Small gain arguments are used to prove asymptotic stability of the points of equilibrium. 
An RG unit is then added to the pre-stabilized system to augment the basin of attraction of the points of equilibrium. 
Numerical simulations and experimental results are provided to demonstrate the effectiveness of the proposed control scheme.

\begin{ack}
	We would like to warmly thank Mr. \emph{Serge Torfs} of our department for his help in the construction of the testbed.
\end{ack}

\bibliographystyle{plain}
\bibliography{settingfiles/Bibliography}

\appendix
\section{Basic properties of system \eqref{roboticsystem1}}\label{app:basicproperties}
\begin{lem}\label{symposdef}
	$M(q)$ is positive-definite and symmetric.
\end{lem}
\begin{pf}
	It is immediate that $M(q)$ is symmetric. 
	$M(q)$ is also positive definite since $M(q)$ is Hermitian and all its eigenvalues are strictly positive.
\end{pf}
\begin{lem}\label{skewsym}
	$[\dot{M}(q)-2C(q,\dot{q})]$ is skew-symmetric.
\end{lem}
\begin{pf}
	The time derivative of $M(q)$ is
	\begin{align}\label{Mdot}
		\dot{M}(q)=\left[\begin{matrix}
				0 & -M_aL\dot{\alpha}\cos\alpha\\
				-M_aL\dot{\alpha}\cos\alpha & 0
		\end{matrix}\right].
	\end{align}
	Then, following from (\ref{Mdot}), we obtain
	\begin{align}
		[\dot{M}(q)-2C(q,\dot{q})]=\left[\begin{matrix}0 & M_aL\dot{\alpha}\cos\alpha\\-M_aL\dot{\alpha}\cos\alpha & 0\end{matrix}\right],
	\end{align}
	which is anti-symmetric and, therefore, skew-symmetric.
\end{pf}

\section{Proof of Lemma \ref{u1coslemma}}\label{u1coslemmaproof}

Using triangular inequality, $\lVert \delta_{\tilde{\theta}} \lVert$ satisfies
\begin{equation}\label{triineq}
	\lVert \delta_{\tilde{\theta}} \lVert \leq \left| f_t \dfrac{\cos\bar{\theta}}{\sin{\bar{\theta}}}\right||\sin\tilde{\theta}| + 2 M_aLg |\cos\alpha|\sin^2(\tilde{\theta}/2).
\end{equation}

First note that $2M_aLg|\cos\alpha|\sin^2(\tilde{\theta}/2)$ is clearly bounded by
\begin{equation}\label{ok1}
	2M_aLg|\cos\alpha|\sin^2(\tilde{\theta}/2)\leq2M_aLg,
\end{equation}
since $|\cos\alpha|\sin^2(\tilde{\theta}/2)\leq 1$ for any $\alpha\in[0,\pi]$, and for any $\tilde{\theta}\in[-\pi,\pi)$. Furthermore, we can say that
\begin{equation}\label{ok2}
	2M_aLg|\cos\alpha|\sin^2(\tilde{\theta}/2)\leq2M_aLg|\tilde{\theta}|,
\end{equation}
since $|\cos\alpha|\leq1$ and $\sin^2(\tilde{\theta}/2)\leq |\tilde{\theta}|$ for any $\alpha\in[0,\pi]$, and for any $\tilde{\theta}\in[\pi,\pi)$.

For what concerns $\left|f_t\dfrac{\cos\bar{\theta}}{\sin{\bar{\theta}}}\right||\sin\tilde{\theta}|$ in (\ref{triineq}), following from (\ref{maptheta}), we have 
\begin{equation}\label{u1cos}
	f_t\dfrac{\cos\bar{\theta}}{\sin{\bar{\theta}}}=\dfrac{f_t}{\tan(\sigma_{\pi/2}(\gamma\arctan(\epsilon f_t)))}.
\end{equation}
For $f_t\not\in[-T_{max}L,T_{max}L]$, due to the saturation $\sigma_{\pi/2}$, we have that $f_t\dfrac{\cos\bar{\theta}}{\sin{\bar{\theta}}}=0$ since $\lim_{\bar{\theta}\to\pm\pi/2}\frac{f_t}{\tan\bar{\theta}}=0$. 

\begin{figure}[!t]\centering{
		\includegraphics[width=0.8\columnwidth]{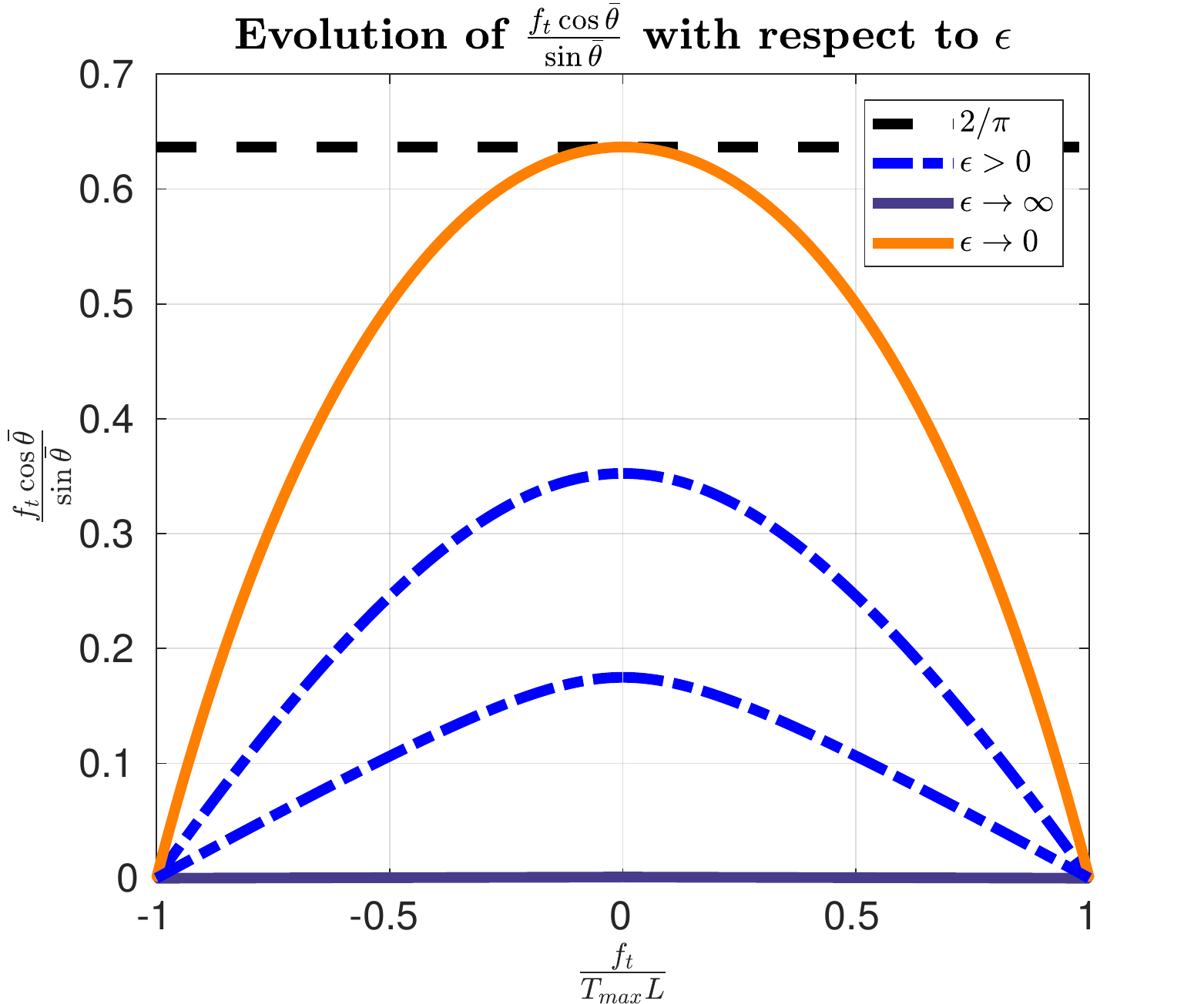}
	\caption{$f_t\dfrac{\cos\bar{\theta}}{\sin\bar{\theta}}$ with $\epsilon\to0$ and $\epsilon\to\infty$.}\label{u1costhetafig}}
\end{figure}

As for $f_t$ restricted to $[-T_{max}L,T_{max}L]$, it is easy to see that (\ref{u1cos}) is continuous as the only potential singularity admits a finite limit, which is
\begin{equation}\label{continuous1}
	\lim_{f_t\to0}\dfrac{f_t}{\tan(\gamma\arctan(\epsilon f_t))}=\dfrac{1}{\gamma\epsilon}.
\end{equation}
Since $f_t\dfrac{\cos\bar{\theta}}{\sin{\bar{\theta}}}$ is continuous and differentiable in the closed interval $f_t\in[-T_{max}L,T_{max}L]$, the possible extrema of (\ref{u1cos}) can be found at the boundaries $f_t=\pm T_{max}L$ and at the stationary points, where $\dfrac{d}{df_t}\left(f_t\dfrac{\cos\bar{\theta}}{\sin{\bar{\theta}}}\right)=0$. 
The only point where $\dfrac{d}{df_t}\left(f_t\dfrac{\cos\bar{\theta}}{\sin{\bar{\theta}}}\right)=0$ is $f_t=0$. 
Therefore, since $f_t\dfrac{\cos\bar{\theta}}{\sin{\bar{\theta}}}\bigg|_{f_t=\pm T_{max}L}=0$ and $f_t\dfrac{\cos\bar{\theta}}{\sin{\bar{\theta}}}\bigg|_{f_t=0}=\dfrac{1}{\gamma\epsilon}$, Eq. (\ref{u1cos}) reaches its maximum when $f_t=0$ (see Fig. \ref{u1costhetafig}). 
In particular, since $\dfrac{1}{\gamma\epsilon}$ is strictly decreasing for $\epsilon\in\mathbb{R}_{>0}$, where $\gamma$ is defined as in (\ref{gamma}) and is a function of $\epsilon$, the maximum of $\dfrac{1}{\gamma\epsilon}$ is $\lim_{\epsilon\to0}\dfrac{1}{\gamma\epsilon}=2/\pi$. 

Consequently, 
\begin{equation}\label{ok3}
	\left|f_t\dfrac{\cos\bar{\theta}}{\sin{\bar{\theta}}}\right||\sin\tilde{\theta}|\leq2/\pi
\end{equation}
for any $\epsilon\in\mathbb{R}_{>0}$, and for any $f_t\in\mathbb{R}$ since $|\sin\tilde{\theta}| \leq 1$ for any $\tilde{\theta}\in[-\pi,\pi)$.
Furthermore, since $|\sin\tilde{\theta}|\leq|\tilde{\theta}|$, we can deduce that
\begin{equation}\label{ok4}
	\left|f_t\dfrac{\cos\bar{\theta}}{\sin\bar{\theta}}\right||\sin\tilde{\theta}|\leq 2/\pi |\tilde{\theta}|
\end{equation}
for any $\epsilon\in\mathbb{R}_{>0}$, any $f_t\in\mathbb{R}$, and any $\tilde{\theta}\in[-\pi,\pi)$.

Following from (\ref{ok1}), (\ref{ok2}), (\ref{ok3}) and (\ref{ok4}), Lemma \ref{u1coslemma} holds true, concluding the proof.

\section{Proof of Lemma \ref{outerloopISS}\label{outerloopISSproof}}

The first part will define the total energy $U_T$, which has useful properties to prove ISS and will be used to define the globally positive definite Lyapunov-candidate-function $V$.
The second step is to prove that $\dot{V}$ is strictly negative with restriction $\lVert \tilde{q} \lVert < \tilde{q}_{max}$ and $|\tilde{\theta}|<\tilde{\theta}_{max}$ whenever the norm $\lVert \tilde{q}, \dot{q} \lVert$ remains outside a ball of radius $\rho|\tilde{\theta}|$.
The last part will be devoted to prove that there exists a finite asymptotic gain $\gamma_{out}$ between $\dot{\bar{\beta}}$ and $\tilde{\beta}$.

Define the total energy $U_T$,
\begin{align}\label{Ut}
	U_T(q_d,\tilde{q},\tilde{\theta})\triangleq\dfrac{1}{2}\tilde{q}^TK_p(\tilde{\theta})\tilde{q}.
\end{align}
Introducing (\ref{Ut}) in (\ref{passage7}), $\tau$ can be rewritten as
\begin{align}\label{controlLaw}
	\tau=\dfrac{\partial{U_T(q_d,\tilde{q},\tilde{\theta})}}{\partial{\tilde{q}}}-K_v(\tilde{\theta})\dot{q}+g(q)+\delta_{\tilde{\theta}}.
\end{align}
In the following, $\tilde{\theta}$ will be restricted to
\begin{equation}\label{thetares}
	\tilde{\theta}\in[-\pi/2+\xi,\pi/2-\xi],
\end{equation}
where $\xi\in\mathbb{R}_{>0}$ is any arbitrarily small positive value.
Denote $\lambda_m\{K_p(\tilde{\theta})\}$ and $\lambda_m\{K_v(\tilde{\theta})\}$ as the smallest eigenvalues of $K_p(\tilde{\theta})$ and of $K_v(\tilde{\theta})$, respectively. 
These eigenvalues are
\begin{align}
	\lambda_m\{K_p(\tilde{\theta})\}&=&k_{p,\alpha}\sin\xi,\label{eigkp}\\
	\lambda_m\{K_v(\tilde{\theta})\}&=&k_{v,\alpha}\sin\xi.
\end{align}
The two following lemmas highlight two important properties of $U_T$ that will be used to prove ISS.

\begin{lem}\label{proposition1}
	\begin{align}
		U_T(q_d,\tilde{q},0)\geq \begin{cases} b \lVert\tilde{q}\lVert^2&\text{, if }\lVert\tilde{q}\lVert<\phi\\
		b\lVert\tilde{q}\lVert& \text{, if }\lVert\tilde{q}\lVert\geq\phi\end{cases},
	\end{align}
	where $b=\dfrac{1}{2}k_{p,\alpha}$ and $\phi=1$.
\end{lem}
\begin{pf}
	The smallest eigenvalue of $K_p(0)$ is $k_{p,\alpha}$. Therefore,
	\begin{align}\label{proofprop3}
		U_T(q_d,\tilde{q},0)=\dfrac{1}{2}\tilde{q}^TK_p(0)\tilde{q}\geq \dfrac{1}{2}k_{p,\alpha}\lVert\tilde{q}\lVert^2,
	\end{align}
	which concludes the proof.
\end{pf}

\begin{lem}\label{proposition2}
	\begin{align}
		\tilde{q}\dfrac{\partial U_T(q_d,\tilde{q},\tilde{\theta})}{\partial \tilde{q}}\geq \begin{cases}b' \lVert\tilde{q}\lVert^2&\text{, if }\lVert\tilde{q}\lVert<\phi'\\b'\lVert\tilde{q}\lVert& \text{, if }\lVert\tilde{q}\lVert\geq\phi'\end{cases},
	\end{align}
	where $b'=k_{p,\alpha}\sin\xi$ and $\phi'=1$.
\end{lem}
\begin{pf}
	Following from (\ref{eigkp}),
	\begin{align}\label{eigkpprout}
		\dfrac{\partial U_T(q_d,\tilde{q},\tilde{\theta})}{\partial \tilde{q}}=
		\tilde{q}^TK_p(\tilde{\theta})\tilde{q}\geq k_{p,\alpha}\sin\xi \lVert\tilde{q}\lVert^2,
	\end{align}
	which concludes the proof.
\end{pf}

At this point, consider the Lyapunov candidate function \eqref{lyapunov}
where
\begin{align}\label{fq}
	f(\tilde{q})\triangleq\dfrac{\tilde{q}}{1+\lVert\tilde{q}\lVert},
\end{align}
and $\gamma_p$ is a positive scalar such that
\begin{equation}\label{alpha}
	\begin{aligned}
		0<\gamma_p <\min\left\{\sqrt{\dfrac{2b}{\lambda_M\{M\}}},\dfrac{2b'}{\lambda_M\{K_v(\tilde{\theta})\}},\right. \\
		\left.\dfrac{\lambda_m\{K_v(\tilde{\theta})\}}{2(k_c+2\lambda_M\{M\})}\right\},
	\end{aligned}
\end{equation}
where $k_c>0$ is any positive scalar satisfying
\begin{align}\label{kccond}
	\lVert C(q,\dot{q})\dot{q}\lVert\leq k_c\lVert\dot{q}\lVert^2.
\end{align}
It is worth noting that $k_c$ in (\ref{kccond}) always exists since it is an inherent property of robotic manipulators \cite{santibanez1997strict}.
The following lemma  proves that (\ref{lyapunov}) under Condition (\ref{alpha}) is a globally positive definite Lyapunov function candidate.
\begin{lem}\label{Vposrad}
	$V$ is radially unbounded and globally positive-definite in $\tilde{q}$ and $\dot{q}$.
\end{lem}
\begin{pf}
	Using the following manipulation,
	\begin{multline}\label{artifice1}
		\dfrac{1}{2}\dot{q}^TM(q)\dot{q}-\gamma_p f(\tilde{q})^TM(q)\dot{q}=\\
		\dfrac{1}{2}[\dot{q}-\gamma_p f(\tilde{q})]^T M(q) [\dot{q}-\gamma_p f(\tilde{q})]\\
		-\dfrac{\gamma_p^2}{2}f(\tilde{q})^TM(q)f(\tilde{q})
	\end{multline}
	and injecting (\ref{artifice1}) in (\ref{lyapunov}), $V$ becomes
	\begin{multline}
		V(\tilde{q},\dot{q})=\dfrac{1}{2}[\dot{q}-\gamma_p f(\tilde{q})]^TM(q)[\dot{q}-\gamma_p f(\tilde{q})]\\
		+U_T(q_d,\tilde{q},0)
		-\dfrac{\gamma_p^2}{2}f(\tilde{q})^TM(q)f(\tilde{q})
	\end{multline}
	Note that, for $\tilde{q}=0$, $f(0)=0$ and $U_T(q_d,0,0)=0$. As a consequence, $V$ becomes
	\begin{equation}
		V(0,\dot{q})=\dfrac{1}{2}\dot{q}^T M(q) \dot{q}.
	\end{equation}
	Since $V(0,\dot{q})>0$ for $\dot{q}\neq 0$ and $V(0,0)=0$, $V(\tilde{q},\dot{q})$ is positive definite for $\tilde{q}=0$. 
	Moreover, since $\lVert\dot{q}\lVert\to\infty\implies V(0,\dot{q})\to\infty$, $V(0,\dot{q})$ is radially unbounded for $\tilde{q}=0$.

	The last step consists in proving that $V(\tilde{q},\dot{q})$ is radially unbounded and positive definite for $\tilde{q}\neq 0$.
	Remark first that the following inequality holds true.
	\begin{equation}
		\begin{aligned}
			\dfrac{\gamma_p^2}{2}f(\tilde{q})^TM(q)f(\tilde{q})=&\dfrac{\gamma_p^2}{2(1+\lVert\tilde{q}\lVert)^2}\tilde{q}^TM(q)\tilde{q}&\\
									    &\leq
			\begin{cases}
				\dfrac{\gamma_p^2}{2}\lambda_M\{M\}\lVert\tilde{q}\lVert^2, & \mbox{if $\lVert\tilde{q}\lVert<\phi$}\\
				\dfrac{\gamma_p^2}{2}\lambda_M\{M\}\lVert\tilde{q}\lVert, & \mbox{if $\lVert\tilde{q}\lVert\geq\phi$},
			\end{cases}
		\end{aligned}
	\end{equation}
	where $\phi=1$. Following from Lemma \ref{proposition1}, as $\phi$ is taken similarly to (\ref{proofprop3}), the following can be deduced.
	\begin{equation}
		\begin{aligned}
			U_T(q_d,\tilde{q},0)-\dfrac{\gamma_p^2}{2}f(\tilde{q})^TM(q)f(\tilde{q})\\
			\geq
			\begin{cases}
				(b-\dfrac{\gamma_p^2}{2}\lambda_M\{M\})\lVert\tilde{q}\lVert^2, & \mbox{if $\lVert\tilde{q}\lVert<\phi$}\\
				(b-\dfrac{\gamma_p^2}{2}\lambda_M\{M\})\lVert\tilde{q}\lVert, & \mbox{if $\lVert\tilde{q}\lVert\geq\phi$},\\
			\end{cases}
		\end{aligned}
	\end{equation}
	making $V$ radially unbounded and positive definite for $\tilde{q}\neq 0$ since $\gamma_p$ satisfies (\ref{alpha}), which concludes the proof.
\end{pf}

At this point, it is worth noting that $\bar{\theta}$ is defined on the reference $\bar{\beta}$ and not on the reference $\bar{\alpha}$, making the error on $\theta$ equal to the error on $\beta$. Therefore, since $\tilde{\theta}=\tilde{\beta}$, it is enough to prove that the controlled system is ISS with respect to $\tilde{\theta}$. To do so, we need to compute the time derivative of V in (\ref{lyapunov}), which is
\begin{equation}\label{proof1}
	\begin{aligned}
		\dot{V}&=\dot{q}^TM(q)\ddot{q}+\dfrac{1}{2}\dot{q}^T\dot{M}(q)\dot{q}-\dot{q}^T\dfrac{\partial U_T(q_d,\tilde{q},0)}{\partial \tilde{q}}\\
		       & -\gamma_p\dot{f}(\tilde{q})^TM(q)\dot{q}-\gamma_p f(\tilde{q})^T\dot{M}(q)\dot{q}-\gamma_p f(\tilde{q})^TM(q)\ddot{q}.
	\end{aligned}
\end{equation}
Substituting $M(q)\ddot{q}$ by its dynamics (\ref{roboticsystem1}) using (\ref{controlLaw}), we obtain
\begin{equation}\label{proof2}
	\begin{aligned}
		\dot{V}&=\dot{q}^T[-C(q,\dot{q})+\dfrac{\partial U_T(q_d,\tilde{q},\tilde{\theta})}{\partial \tilde{q}}-K_v(\tilde{\theta})\dot{q}+\delta_{\tilde{\theta}}]\\
		       &+\dfrac{1}{2}\dot{q}^T\dot{M}(q)\dot{q}-\dot{q}^T\dfrac{\partial U_T(q_d,\tilde{q},0)}{\partial \tilde{q}}\\
		       & -\gamma_p\dot{f}(\tilde{q})^TM(q)\dot{q}-\gamma_p f(\tilde{q})^T\dot{M}(q)\dot{q}\\
		       & -\gamma_p f(\tilde{q})^T[-C(q,\dot{q})\dot{q}+\dfrac{\partial U_T(q_d,\tilde{q},\tilde{\theta})}{\partial \tilde{q}}-K_v(\tilde{\theta})\dot{q}+\delta_{\tilde{\theta}}].
	\end{aligned}
\end{equation}
Then, using Lemma \ref{skewsym}, (\ref{proof2}) becomes
\begin{equation}\label{proof4}
	\begin{aligned}
		\dot{V}&=\dot{q}^T[\dfrac{\partial U_T(q_d,\tilde{q},\tilde{\theta})}{\partial \tilde{q}}-K_v(\tilde{\theta})\dot{q}+\delta_{\tilde{\theta}}]\\
		       &-\dot{q}^T\dfrac{\partial U_T(q_d,\tilde{q},0)}{\partial \tilde{q}}-\gamma_p\dot{f}(\tilde{q})^TM(q)\dot{q}-\gamma_p f(\tilde{q})^T\dot{M}\dot{q}\\
		       & -\gamma_p f(\tilde{q})^T[-C(q,\dot{q})\dot{q}+\dfrac{\partial U_T(q_d,\tilde{q},\tilde{\theta})}{\partial \tilde{q}}-K_v(\tilde{\theta})\dot{q}+\delta_{\tilde{\theta}}].
	\end{aligned}
\end{equation}
Moreover, since Lemma \ref{skewsym} holds true, $\dot{M}=C(q,\dot{q})+C(q,\dot{q})^T$ also holds true. Then, (\ref{proof4}) becomes
\begin{equation}\label{proof4bis}
	\begin{aligned}
		\dot{V}&=\dot{q}^T[\dfrac{\partial U_T(q_d,\tilde{q},\tilde{\theta})}{\partial \tilde{q}}-K_v(\tilde{\theta})\dot{q}+\delta_{\tilde{\theta}}]\\
		       &-\dot{q}^T\dfrac{\partial U_T(q_d,\tilde{q},0)}{\partial \tilde{q}}-\gamma_p\dot{f}(\tilde{q})^TM(q)\dot{q}\\
		       & -\gamma_p f(\tilde{q})^T[C(q,\dot{q})^T\dot{q}+\dfrac{\partial U_T(q_d,\tilde{q},\tilde{\theta})}{\partial \tilde{q}}-K_v(\tilde{\theta})\dot{q}+\delta_{\tilde{\theta}}].
	\end{aligned}
\end{equation}
Substituting the expression of $U_T$ in (\ref{proof4bis}), we obtain
\begin{equation}\label{proof5}
	\begin{aligned}
		\dot{V}&=-\dot{q}^TK_v(\tilde{\theta})\dot{q}+\dot{q}^T[K_p(\tilde{\theta})-K_p(0)]\tilde{q}-\gamma_p\dot{f}(\tilde{q})^TM(q)\dot{q}\\
		       &-\gamma_p f(\tilde{q})^T[C(q,\dot{q})^T\dot{q}+K_p(\tilde{\theta})\tilde{q}-K_v(\tilde{\theta})\dot{q}]\\
		       & +[\dot{q}-\gamma_p f(\tilde{q})]^T\delta_{\tilde{\theta}}.
	\end{aligned}
\end{equation}

In order to bound $\dot{V}$, note that the following inequalities hold true \cite{santibanez1997strict},
\begin{align}\label{fqcond}
	f(\tilde{q})\leq&  1,\\
	\dot{f}(\tilde{q})\leq&  2 \lVert \dot{q} \lVert ,\label{fqcond2}
\end{align}
and that the three following inequalities can be deduced using (\ref{kccond}), (\ref{fqcond}) and (\ref{fqcond2}).
\begin{align}\label{proof6}
	-\dot{q}^TK_v(\tilde{\theta})\dot{q}&\leq & -\dfrac{1}{2}\dot{q}^TK_v(\tilde{\theta})\dot{q}\\
					    &&-\dfrac{1}{2}\lambda_m\{K_v(\tilde{\theta})\} \lVert \dot{q} \lVert ^2,\nonumber \\
	\label{proof6bis}-\gamma_p \dot{f}(\tilde{q})^TM(q)\dot{q} & \leq & 2\gamma_p\lambda_M\{M\} \lVert \dot{q} \lVert ^2, \\
	\label{proof6ter}-\gamma_p f(\tilde{q})^TC(q,\dot{q})^T\dot{q} &\leq & \gamma_p k_c  \lVert \dot{q} \lVert ^2.
\end{align}
Following from (\ref{proof6})--(\ref{proof6ter}), (\ref{proof5}) can be bounded by
\begin{equation}\label{proof7}
	\begin{aligned}
		\dot{V}&\leq-\dfrac{1}{2}\dot{q}^TK_v(\tilde{\theta})\dot{q}-[\dfrac{1}{2}\lambda_m\{K_v(\tilde{\theta})\}-2\gamma_p\lambda_M\{M\}\\
		       &-\gamma_p k_c] \lVert \dot{q} \lVert ^2+\dot{q}^T[K_p(\tilde{\theta})-K_p(0)]\tilde{q}\\
		       &-\gamma_p f(\tilde{q})^T[K_p(\tilde{\theta})\tilde{q}-K_v(\tilde{\theta})\dot{q}]+[\dot{q}-\gamma_p f(\tilde{q})]^T\delta_{\tilde{\theta}}.
	\end{aligned}
\end{equation}

Introducing the artifice
\begin{equation}\label{proof8}
	\begin{aligned}
		-\dfrac{1}{2}\dot{q}^TK_v(\tilde{\theta})\dot{q}+&\gamma_p f(\tilde{q})^TK_v(\tilde{\theta})\dot{q}=\\&-\dfrac{1}{2}[\dot{q}-\gamma_p f(\tilde{q})]^TK_v(\tilde{\theta})[\dot{q}-\gamma_p f(\tilde{q})]\\
								 &+\dfrac{\gamma_p^2}{2} f(\tilde{q})^T K_v(\tilde{\theta}) f(\tilde{q}),
	\end{aligned}
\end{equation}
we can rewrite (\ref{proof7}) using (\ref{proof8}) as
\begin{equation}\label{proof9}
	\begin{aligned}
		\dot{V}&\leq-\dfrac{1}{2}[\dot{q}-\gamma_p f(\tilde{q})]^TK_v(\tilde{\theta})[\dot{q}-\gamma_p f(\tilde{q})]\\
		       &-[\dfrac{1}{2}\lambda_m\{K_v(\tilde{\theta})\}-2\gamma_p\lambda_M\{M\}-\gamma_p k_c] \lVert \dot{q} \lVert ^2\\
		       &+\dot{q}^T[K_p(\tilde{\theta})-K_p(0)]\tilde{q}\\
		       &-\gamma_p f(\tilde{q})^T[K_p(\tilde{\theta})\tilde{q}-\dfrac{\gamma_p K_v(\tilde{\theta})f(\tilde{q})}{2}]+[\dot{q}-\gamma_p f(\tilde{q})]^T\delta_{\tilde{\theta}}.
	\end{aligned}
\end{equation}
Substituting (\ref{fq}) in (\ref{proof9}), we obtain
\begin{equation}\label{proof10}
	\begin{aligned}
		\dot{V}&\leq-\dfrac{1}{2}[\dot{q}-\gamma_p f(\tilde{q})]^TK_v(\tilde{\theta})[\dot{q}-\gamma_p f(\tilde{q})]\\
		       &-[\dfrac{1}{2}\lambda_m\{K_v(\tilde{\theta})\}-2\gamma_p\lambda_M\{M\}-\gamma_p k_c] \lVert \dot{q} \lVert ^2\\
		       &+\dot{q}^T[K_p(\tilde{\theta})-K_p(0)]\tilde{q}\\
		       &-\dfrac{\gamma_p}{1+ \lVert \tilde{q} \lVert }[\tilde{q}^TK_p(\tilde{\theta})\tilde{q}-\dfrac{\gamma_p \tilde{q}^T K_v(\tilde{\theta})\tilde{q}}{2(1+ \lVert \tilde{q} \lVert )}]\\
		       &+[\dot{q}-\gamma_p f(\tilde{q})]^T\delta_{\tilde{\theta}}.
	\end{aligned}
\end{equation}

At this point, to prove ISS with restriction, it is enough to prove that $\dot{V}<0$ whenever $\lVert \tilde{q},\dot{q} \lVert$ remains outside a certain ball of radius $\rho|\tilde{\theta}|$ with restriction $\lVert \tilde{q} \lVert < \tilde{q}_{max}$ and $|\tilde{\theta}|<\tilde{\theta}_{max}$. To do so, we will use the following lemma.
\begin{lem}\label{noveltylemma}
	The following inequality 
	\begin{equation}\label{proof15}
		\begin{aligned}
			[\dfrac{1}{2}\lambda_m\{K_v(\tilde{\theta})\}-2\gamma_p\lambda_M\{M\}-\gamma_p k_c] \lVert \dot{q} \lVert ^2&\\
			+\dot{q}^T[K_p(0)-K_p(\tilde{\theta})]\tilde{q}&\\
			+\dfrac{\gamma_p}{1+ \lVert \tilde{q} \lVert }[\tilde{q}^TK_p(\tilde{\theta})\tilde{q}-\dfrac{\gamma_p \tilde{q}^T K_v(\tilde{\theta})\tilde{q}}{2(1+ \lVert \tilde{q} \lVert )}]& \geq \mu \left\lVert\begin{matrix}
					\tilde{q} \\ \dot{q}
				\end{matrix}\right\lVert^2,
			\end{aligned}
		\end{equation}
		holds true with restrictions $|\tilde{\theta}|<\tilde{\theta}_{max}$ and $\lVert \tilde{q} \lVert < \tilde{q}_{max}$.
		The restrictions on $\tilde{q}_{max}$ must satisfy $\tilde{q}_{max}\triangleq \min\{\tilde{q}_{max,1},\tilde{q}_{max,2}\}$, where
		\begin{align}
			\tilde{q}_{max,1}=&\dfrac{2 \gamma_p k_{p,x}}{2 \mu + \gamma_p^2 k_{v,x}} -1,\label{condqmax1}\\
			\tilde{q}_{max,2}=&\dfrac{2\gamma_p k_{p,\alpha}}{2\mu + \gamma_p^2 k_{v,\alpha}}-1,\label{condqmax2}
		\end{align}
		and $\mu$ is any constant satisfying
		\begin{align}\label{mucond}
			0 < \mu <
			\min\left\{\dfrac{1}{2}\lambda_m\{K_v(\tilde{\theta})\}-2 \gamma_p \lambda_M\{M\}- \gamma_p k_c,\right.\\
			\left.\gamma_p k_{p,x}-\dfrac{\gamma_p^2 k_{v,x}}{2},\gamma_p k_{p,\alpha}-\dfrac{\gamma_p^2 k_{v,\alpha}}{2}\right\}.
		\end{align}
		The restrictions on $\tilde{\theta}_{max}$ must satisfy $\tilde{\theta}_{max}\triangleq\min\{\tilde{\theta}_{max,1},\allowbreak\tilde{\theta}_{max,2}\}$, where
		\begin{align}
			\tilde{\theta}_{max,1}=&\left|\arccos\left\{\dfrac{k_{p,\alpha}+\mu}{\frac{\gamma_p k_{p,\alpha}}{1+\tilde{q}_{max,2}} - \frac{\gamma_p^2 k_{v,\alpha}}{2}+k_{p,\alpha}}\right\}\right|,\label{thetamax1}
		\end{align}
		\begin{align}
			\tilde{\theta}_{max,2}=
		&\left|\arccos\left\{\dfrac{-\frac{1}{2}\lambda_m\{K_v(\tilde{\theta})\}+2\gamma_p\lambda_M\{M\}+\gamma_p k_c}{k_{p,\alpha}}\right.\right.\nonumber\\
		&\left.\left.+\dfrac{\mu+k_{p,\alpha}}{k_{p,\alpha}}\right\}\right|.\label{thetamax2}
		\end{align}
	\end{lem}

	\begin{pf}
		To analyze (\ref{proof15}), let us rewrite the expression in the matrix form
		\begin{multline}\label{matrixsecond}
			\begin{aligned}
				[\begin{matrix} \tilde{x} & \tilde{\alpha} & \dot{x} & \dot{\alpha}\end{matrix}]
				\left[
					\begin{matrix}
						p_{11} & 0 & 0 & 0\\
						0 & p_{22} & 0 & p_{24}\\
						0 & 0 & p_{33} & 0\\
						0 & p_{42} & 0 & p_{44}
					\end{matrix}
					\right]\left[\begin{matrix}
						\tilde{x}\\ \tilde{\alpha}\\ \dot{x}\\ \dot{\alpha}
				\end{matrix}\right]\\
				\geq 
				[\begin{matrix} \tilde{x} & \tilde{\alpha} & \dot{x} & \dot{\alpha}\end{matrix}]
				\mu I
				\left[\begin{matrix}
						\tilde{x}\\ \tilde{\alpha}\\ \dot{x}\\ \dot{\alpha}
				\end{matrix}\right]
			\end{aligned}
		\end{multline}
		where $p_{11}\triangleq\dfrac{\gamma_p k_{p,x}}{1+ \lVert \tilde{q} \lVert } - \dfrac{\gamma_p^2 k_{v,x}}{2(1+ \lVert \tilde{q} \lVert )^2}$, $p_{22}\triangleq\dfrac{\gamma_p k_{p,\alpha}\cos\tilde{\theta}}{1+ \lVert \tilde{q} \lVert } - \dfrac{\gamma_p^2 k_{v,\alpha}\cos\tilde{\theta}}{2(1+ \lVert \tilde{q} \lVert )^2}$, $p_{33}=p_{44}\triangleq\dfrac{1}{2}\lambda_m\{K_v(\tilde{\theta})\}-2\gamma_p\lambda_M\{M\}-\gamma_p k_c$, $p_{42}=p_{24}\triangleq k_{p,\alpha}(1-\cos\tilde{\theta})$ and $I$ is the identity matrix.

		Equivalently to (\ref{matrixsecond}), we can write
		\begin{multline}\label{matrixthird}
			\begin{aligned}
				[\begin{matrix} \tilde{x} & \tilde{\alpha} & \dot{x} & \dot{\alpha}\end{matrix}]
				\left[
					\begin{matrix}
						p'_{11} & 0 & 0 & 0\\
						0 & p'_{22} & 0 & p_{24}\\
						0 & 0 & p'_{33} & 0\\
						0 & p_{42} & 0 & p'_{44}
					\end{matrix}
					\right]\left[\begin{matrix}
						\tilde{x}\\ \tilde{\alpha}\\ \dot{x}\\ \dot{\alpha}
				\end{matrix}\right]
				\geq 
				0,
			\end{aligned}
		\end{multline}
		where $p'_{11}=p_{11}-\mu$, $p'_{22}=p_{22}-\mu$, $p'_{33}=p_{33}-\mu$ and $p'_{44}=p_{44}-\mu$.

		Let $P$ be the square matrix in (\ref{matrixthird}). 
		It is important to note that $P$ is Hermitian since $P$ is symmetric and all its entries are real numbers.
		Moreover, note that $p_{24},p_{42}>0$ since $\tilde{\theta}$ is restricted to (\ref{thetares}).
		Accordingly, it is sufficient to prove that $P$ is strictly diagonally dominant to ensure (\ref{proof15}). 

		In particular, we have to prove
		\begin{align}
			p'_{11} > & 0, \label{p11} \\
			p'_{22} > & p_{24}, \label{p22} \\
			p'_{33} > & 0,  \label{p33} \\
			p'_{44} > & p_{42}. \label{p44}
		\end{align}

		First, let us rewrite (\ref{p11}) as
		\begin{align}\label{p11ex}
			\dfrac{\gamma_p k_{p,x}}{1+\lVert \tilde{q}\lVert}-\dfrac{\gamma_p ^2 k_{v,x}}{2(1+\lVert \tilde{q} \lVert ) ^2} > \mu.
		\end{align}
		Because of the choice of $\gamma_p$ in (\ref{alpha}), the left-hand side of (\ref{p11ex}) is strictly positive.
		Indeed, we can remark that
		\begin{equation}
			-\dfrac{\gamma_p \tilde{q}^T K_v(\tilde{\theta}) \tilde{q}}{2(1+\lVert \tilde{q} \lVert)}\geq
			\begin{aligned}
				\begin{cases}
					-\lambda_M\{K_v(\tilde{\theta})\}/2\lVert \tilde{q} \lVert ^2, & \mbox{if $\lVert \tilde{q} \lVert < \phi'$}\\
					-\lambda_M\{K_v(\tilde{\theta})\}/2\lVert \tilde{q} \lVert , & \mbox{if $\lVert \tilde{q} \lVert \geq \phi'$},\\
				\end{cases}
			\end{aligned}
		\end{equation}
		for $\phi'=1$. Following from Lemma \ref{proposition2} and from restriction (\ref{alpha}), since $\phi'$ is the same as in (\ref{eigkpprout}), we can write
		\begin{equation}
			\tilde{q}^T K_P(\tilde{\theta}) \tilde{q} \geq b' \lVert \tilde{q} \lVert ^2 > \dfrac{\gamma_p}{2} \lambda_M\{K_v\}\lVert \tilde{q} \lVert ^2 \geq\dfrac{\gamma_p \tilde{q}^T K_v(\tilde{\theta}) \tilde{q}}{2(1+\lVert \tilde{q} \lVert)},
		\end{equation}
		for all $0<\lVert \tilde{q} \lVert <\phi'$, and
		\begin{equation}
			\tilde{q}^T K_P(\tilde{\theta}) \tilde{q} \geq b' \lVert \tilde{q} \lVert  > \dfrac{\gamma_p}{2} \lambda_M\{K_v\}\lVert \tilde{q} \lVert \geq\dfrac{\gamma_p \tilde{q}^T K_v(\tilde{\theta}) \tilde{q}}{2(1+\lVert \tilde{q} \lVert)}
		\end{equation}
		for all $\lVert \tilde{q} \lVert \geq \phi'$.

		Consequently, since $\gamma_p$ and $\mu$ are fixed (cf. (\ref{alpha}) and (\ref{mucond}), respectively), the norm $\lVert \tilde{q} \lVert$ must be restricted to satisfy (\ref{p11ex}).
		Indeed, we need to restrict $\lVert \tilde{q} \lVert$ to $\tilde{q}_{max,1}$ considering the worst case scenario of (\ref{p11ex}), which is
		\begin{align}\label{tropfaim}
			\dfrac{\gamma_p k_{p,x}}{1+\tilde{q}_{max,1}}-\dfrac{\gamma_p^2 k_{v,x}}{2}>\mu.
		\end{align}
		Remark that the left-hand side of (\ref{tropfaim}) is a decreasing function in $\tilde{q}_{max,1}$ and is positive when $\tilde{q}_{max,1}=0$ because of the choice of $\gamma_p$ in (\ref{alpha}). Moreover, a solution exists for (\ref{tropfaim}) since $\mu$ satisfies (\ref{mucond}).
		Therefore, to ensure (\ref{p11ex}), we must satisfy
		\begin{align}
			\tilde{q}_{max,1}<\dfrac{2 \gamma_p k_{p,x}}{2 \mu + \gamma_p^2 k_{v,x}} -1.
		\end{align}

		The next step is to verify Condition (\ref{p22}), which can be rewritten as
		\begin{align}\label{diagdom}
			\dfrac{\gamma_p k_{p,\alpha}\cos\tilde{\theta}}{1+ \lVert \tilde{q} \lVert } - \dfrac{\gamma_p^2 k_{v,\alpha}\cos\tilde{\theta}}{2(1+ \lVert \tilde{q} \lVert )^2} &> \mu + k_{p,\alpha}(1-\cos\tilde{\theta})
		\end{align}
		Following the same arguments as (\ref{p11ex}), the left-hand side  of (\ref{diagdom}) is strictly positive due to the choice of $\gamma_p$ in (\ref{alpha}). 
		Considering the worst case scenario of (\ref{diagdom}), which is
		\begin{align}\label{diagdom2}
			\dfrac{\gamma_p k_{p,\alpha}\cos\tilde{\theta}}{1+\tilde{q}_{max,2}} - \dfrac{\gamma_p^2 k_{v,\alpha}\cos\tilde{\theta}}{2} &> \mu + k_{p,\alpha}(1-\cos\tilde{\theta}),
		\end{align}
		we can remark that the left-hand side of (\ref{diagdom2}) is a decreasing function in $\tilde{q}_{max,2}$ and is positive when $\tilde{q}_{max,2}=0$ because of (\ref{alpha}). Moreover, following from the choice of $\mu$ in (\ref{mucond}), we can ensure that (\ref{diagdom2}) is solvable. 
		Note also that the right-hand side of (\ref{diagdom2}) is positive since $\mu>0$ and $0<\cos\tilde{\theta}\leq 1$ for any $\tilde{\theta}\in[-\pi/2+\xi,\pi/2-\xi]$. 

		We can deduce that Condition (\ref{diagdom2}) additionally restricts $\lVert \tilde{q} \lVert$ to $\tilde{q}_{max,2}$. Indeed, rewriting (\ref{diagdom2}) as
		\begin{equation}\label{cosstrange}
			\cos\tilde{\theta}>\dfrac{k_{p,\alpha}+\mu}{k_{p,\alpha}+\frac{\gamma_p k_{p,\alpha}}{1+\tilde{q}_{max,2}}-\frac{\gamma_p^2 k_{v,\alpha}}{2}},
		\end{equation}
		it is worth to remark that (\ref{cosstrange}) is solvable if and only if the right hand-side of (\ref{cosstrange}) is lower than the unity.
		This condition is satisfied only if
		\begin{equation}
			\mu < \dfrac{\gamma_p k_{p,\alpha}}{1+\tilde{q}_{max,2}}-\dfrac{\gamma_p^2 k_{v,\alpha}}{2},
		\end{equation}
		which can be rewritten as
		\begin{equation}
			\tilde{q}_{max,2}<\dfrac{2\gamma_p k_{p,\alpha}}{2\mu + \gamma_p^2 k_{v,\alpha}}-1.
		\end{equation}

		Consequently, to satisfy (\ref{cosstrange}), $\tilde{\theta}$ must be restricted to $| \tilde{\theta} | <\tilde{\theta}_{max,1}$, where $\tilde{\theta}_{max,1}$ is expressed in (\ref{thetamax1}).

		For what concerns Condition (\ref{p33}), following from Definition (\ref{mucond}), we can immediately deduce that this condition is satisfied since
		\begin{align}
			p'_{33}=\dfrac{1}{2}\lambda_m\{K_v(\tilde{\theta})\}-2\gamma_p \lambda_M\{M\}-\gamma_p k_c -\mu > 0.
		\end{align}

		The final step is to rewrite (\ref{p44}) as
		\begin{align}\label{lastdiag}
			\dfrac{1}{2}\lambda_m\{K_v(\tilde{\theta})\}-2\gamma_p\lambda_M\{M\}-\gamma_p k_c - \mu > k_{p,\alpha}(1-\cos\tilde{\theta}).
		\end{align}
		Remark that the left hand-side of (\ref{lastdiag}) is strictly positive because of the choice of $\mu$ in (\ref{mucond}).

		There are two cases.
		\begin{enumerate}[(i)]
			\item If $\dfrac{1}{2}\lambda_m\{K_v(\tilde{\theta})\}-2\gamma_p\lambda_M\{M\}-\gamma_p k_c-\mu>k_{p,\alpha}$, then (\ref{lastdiag}) is always satisfied for $\tilde{\theta}\in[-\pi/2+\xi,\pi/2-\xi]$.
			\item Otherwise, $\tilde{\theta}$ is restricted to $|\tilde{\theta}|<\tilde{\theta}_{max,2}$, where $\tilde{\theta}_{max,2}$ is expressed in (\ref{thetamax2}).

		\end{enumerate}

		To summarize, to satisfy (\ref{p11}) and (\ref{p22}), $\lVert \tilde{q} \lVert$ must be restricted to
		\begin{equation}\label{qmaxfinal}
			\lVert \tilde{q} \lVert < \min\{\tilde{q}_{max,1},\tilde{q}_{max,2}\},
		\end{equation}
		and to satisfy (\ref{p22}) and (\ref{p44}), $\tilde{\theta}$ must be restricted to
		\begin{equation}\label{thetafinal}
			|\tilde{\theta}| < \min\{\tilde{\theta}_{max,1},\tilde{\theta}_{max,2}\},
		\end{equation}
		which concludes the proof.
	\end{pf}

	Using Lemma \ref{noveltylemma}, (\ref{proof10}) can be bounded by
	\begin{multline}
		\dot{V}\leq -\dfrac{1}{2}[\dot{q}-\gamma_p f(\tilde{q})]^T K_v(\tilde{\theta}) [\dot{q}-\gamma_p f(\tilde{q})] \\
		- \mu \lVert \tilde{q} , \dot{q} \lVert ^2 + [\dot{q} - \gamma_p f(\tilde{q})] \delta_{\tilde{\theta}},
	\end{multline}
	where $\mu$ must satisfy (\ref{mucond}). Since
	\begin{equation}
		\lVert f(\tilde{q}) \lVert \leq \lVert \tilde{q} \lVert,
	\end{equation}
	using the triangular inequality, we can write
	\begin{multline}\label{finalstepproof}
		\dot{V} \leq -\dfrac{1}{2}[\dot{q}-\gamma_p f(\tilde{q})]^T K_v(\tilde{\theta}) [\dot{q}-\gamma_p f(\tilde{q})]\\
		-\mu \lVert \tilde{q} \lVert ^2 - \mu \lVert \dot{q} \lVert ^2 + \lVert \dot{q} \lVert \lVert \delta_{\tilde{\theta}} \lVert + \gamma_p \lVert \tilde{q} \lVert \lVert \delta_{\tilde{\theta}} \lVert.
	\end{multline}
	The first term on the right-hand side of (\ref{finalstepproof}) is negative semi-definite since this term only cancels out when $\dot{q}=\gamma_p f(\tilde{q})$. 
	As a consequence, $\lVert \tilde{q} \lVert$ and $\lVert \dot{q} \lVert$ must satisfy
	\begin{align}
		\lVert \tilde{q} \lVert > & \dfrac{\gamma_p \lVert \delta_{\tilde{\theta}} \lVert}{\mu},\\
		\lVert \dot{q} \lVert > & \gamma_p \lVert \delta_{\tilde{\theta}} \lVert.
	\end{align}
	Therefore, $\dot{V}<0$ when $|\tilde{\theta}|$ and $\lVert \tilde{q} \lVert$ are restricted to $\tilde{\theta}_{max}$ and $\tilde{q}_{max}$, respectively, and if $\lVert \tilde{q}, \dot{q} \lVert$ remains outside the ball
	\begin{equation}\label{ISSfinal}
		\lVert \tilde{q}, \dot{q} \lVert > \dfrac{\gamma_p \sqrt{1+\mu^2}}{\mu}\lVert \delta_{\tilde{\theta}} \lVert.
	\end{equation}
	Note that, since $\tilde{\theta}=\tilde{\beta}$, following from Lemma \ref{u1coslemma} and (\ref{ISSfinal}), $\dot{V}<0$ when $|\tilde{\theta}|<\tilde{\theta}_{max}$ and $\lVert \tilde{q} \lVert < \tilde{q}_{max}$ if
	\begin{equation}\label{ISSfinal2}
		\lVert \tilde{q},\dot{q} \lVert > \rho|\tilde{\beta}|,
	\end{equation}
	where $\rho\triangleq\dfrac{\gamma_p \sqrt{1+\mu^2}(2/\pi+2M_aLg)}{\mu}$, concluding the proof for ISS with respect to $\tilde{\beta}$.

	The final step is to compute the asymptotic gain $\gamma_{out}$ between $\tilde{\beta}$ and $\dot{\bar{\beta}}$.
	Using triangular inequality, $|\dot{\bar{\beta}}|$ is bounded by
	\begin{equation}\label{betabardotbound1}
		|\dot{\bar{\beta}}|\leq|\dot{\bar{\theta}}|+|\dot{\alpha}|.
	\end{equation}
	Then, computing the time derivative of (\ref{maptheta}) using (\ref{ftexplicit}) for $f_t\in[-T_{max}L,T_{max}L]$, we obtain
	\begin{equation}
		\dot{\bar{\theta}} =\dfrac{\epsilon(k_{p,\alpha}\dot{\alpha}-k_{v,\alpha}\ddot{\alpha}-M_aLg\sin\alpha\dot{\alpha})}{\epsilon^2(k_{p,\alpha}\tilde{\alpha}-k_{v,\alpha}\dot{\alpha}+M_aLg\cos\alpha)^2+1},
	\end{equation}
	whose norm can be bounded by
	\begin{equation}
		|\dot{\bar{\theta}}|\leq \epsilon(k_{p,\alpha}|\dot{\alpha}|+k_{v,\alpha}|\ddot{\alpha}|+M_aLg|\sin\alpha||\dot{\alpha}|).
	\end{equation}

	Since $|\dot{\alpha}|\leq\lVert\dot{q}\lVert$, $|\ddot{\alpha}|\leq\lVert\ddot{q}\lVert$ and $|\sin\alpha|\leq1$, it is possible to build a direct relationship between the norm $|\dot{\bar{\theta}}|$ and the norm of the states of (\ref{roboticsystem1}) as
	\begin{equation}\label{thetabardotbound}
		|\dot{\bar{\theta}}|\leq\epsilon[(k_{p,\alpha}+M_aLg)\lVert\dot{q}\lVert+k_{v,\alpha}\lVert\ddot{q}\lVert].
	\end{equation}
	Injecting (\ref{thetabardotbound}) in (\ref{betabardotbound1}) and using the same previous argument $|\dot{\alpha}|\leq\lVert\dot{q}\lVert$, we obtain
	\begin{equation}\label{betabardotbound2}
		|\dot{\bar{\beta}}|\leq [\epsilon(k_{p,\alpha}+M_aLg)+1]\lVert\dot{q}\lVert+\epsilon k_{v,\alpha}\lVert\ddot{q}\lVert.
	\end{equation}

	In order to express (\ref{betabardotbound2}) in terms of the states $\tilde{q}$ and $\dot{q}$, let us rewrite $\ddot{q}$ using (\ref{roboticsystem1}) and (\ref{passage7}) as
	\begin{equation}\label{ddotqex}
		\ddot{q}=M^{-1}(q)[-C(q,\dot{q})\dot{q}+K_p(\tilde{\theta})\tilde{q}-K_v(\tilde{\theta})\dot{q}+\delta_{\tilde{\theta}}].
	\end{equation}
	Using (\ref{kccond}), the norm of (\ref{ddotqex}) can be bounded by
	\begin{multline}\label{ddotqbound}
		\lVert \ddot{q} \lVert \leq \dfrac{1}{\lambda_m\{M\}}[k_c\lVert \dot{q} \lVert + \lambda_M\{K_p(\tilde{\theta})\}\lVert \tilde{q} \lVert \\
		+ \lambda_M\{K_v(\tilde{\theta})\}\lVert \dot{q} \lVert + \lVert \delta_{\tilde{\theta}} \lVert],
	\end{multline}
	where $\lambda_m\{M\}$ is the minimal eigenvalue of $M(q)$ and $\lambda_M\{K_p(\tilde{\theta})\}\triangleq k_{p,x}$ the maximal eigenvalue of $K_p(\tilde{\theta})$.
	Injecting (\ref{ddotqbound}) in (\ref{betabardotbound2}), we obtain
	\begin{equation}\label{betabardotbound3}
		|\dot{\bar{\beta}}|\leq \lambda_{\dot{q}}\lVert \dot{q} \lVert +\lambda_{\tilde{q}}\lVert \tilde{q} \lVert + \lambda_\delta\lVert \delta_{\tilde{\theta}} \lVert,
	\end{equation}
	where $\lambda_{\dot{q}}\triangleq\epsilon(k_{p,\alpha}+M_aLg)+1+\dfrac{\epsilon k_{v,\alpha}}{\lambda_m\{M\}}(k_c+\lambda_M\{K_v(\tilde{\theta})\})$, $\lambda_{\tilde{q}}\triangleq\dfrac{\epsilon k_{v,\alpha} \lambda_M\{K_p(\tilde{\theta})\} }{\lambda_m\{M\}}$ and $\lambda_\delta\triangleq\dfrac{\epsilon k_{v,\alpha}}{\lambda_m\{M\}}$.

	Since $\lVert \tilde{q},\dot{q} \lVert \geq \lVert \tilde{q} \lVert$, $\lVert \tilde{q},\dot{q} \lVert \geq \lVert \dot{q} \lVert$ and $\dot{V}<0$ when $|\tilde{\theta}|<\tilde{\theta}_{max}$ and $\lVert \tilde{q} \lVert < \tilde{q}_{max} $ if (\ref{ISSfinal2}) holds true, we can say that, eventually, $\lVert \tilde{q} \lVert$ and $\lVert \dot{q} \lVert$ will be contained in a ball of radius $\rho|\tilde{\theta}|$.
	In this case, (\ref{betabardotbound3}) becomes
	\begin{equation}
		|\dot{\bar{\beta}}|\leq(\lambda_{\dot{q}}+\lambda_{\tilde{q}})\rho|\tilde{\theta}|+\lambda_\delta\lVert\delta_{\tilde{\theta}}\lVert.
	\end{equation}

	Finally, since $\tilde{\theta}=\tilde{\beta}$, using Lemma \ref{u1coslemma}, we can write
	\begin{equation}
		|\dot{\bar{\beta}}|\leq \gamma_{out}|\tilde{\beta}|,
	\end{equation}
	where
	\begin{equation}\label{gammaout}
		\gamma_{out}\triangleq[(\lambda_{\dot{q}}+\lambda_{\tilde{q}})\rho+\lambda_\delta(2/\pi+2M_aLg)],
	\end{equation}
	which concludes the proof of the existence of a finite asymptotic gain $\gamma_{out}$ between $\tilde{\beta}$ and $\dot{\bar{\beta}}$.

	\section{Proof of Proposition \ref{propositionu1new}}\label{proofu1new}

	First let $f_t$ denote the reference for the tangential force that is requested by the system. 
	Let $f_{old}$ and $f_{new}$ be the actual forces delivered to the object using the control law (\ref{mapu1}) and (\ref{u1new}), respectively, which are
	\begin{equation}\label{fnewfold}
		\begin{aligned}
			\begin{cases}
				f_{old}&\triangleq\dfrac{f_t\sin\theta}{\sin\bar{\theta}}\\&=\dfrac{f_t\sin(\bar{\theta}+\tilde{\theta})}{\sin{\bar{\theta}}}\\
				f_{new}&\triangleq\sigma_{0,T_{max}L}\left(\dfrac{f_t}{\sin\theta}\right)\sin\theta.
			\end{cases}
		\end{aligned}
	\end{equation}

\begin{figure}[!t]\centering{
			\includegraphics[width=0.8\columnwidth]{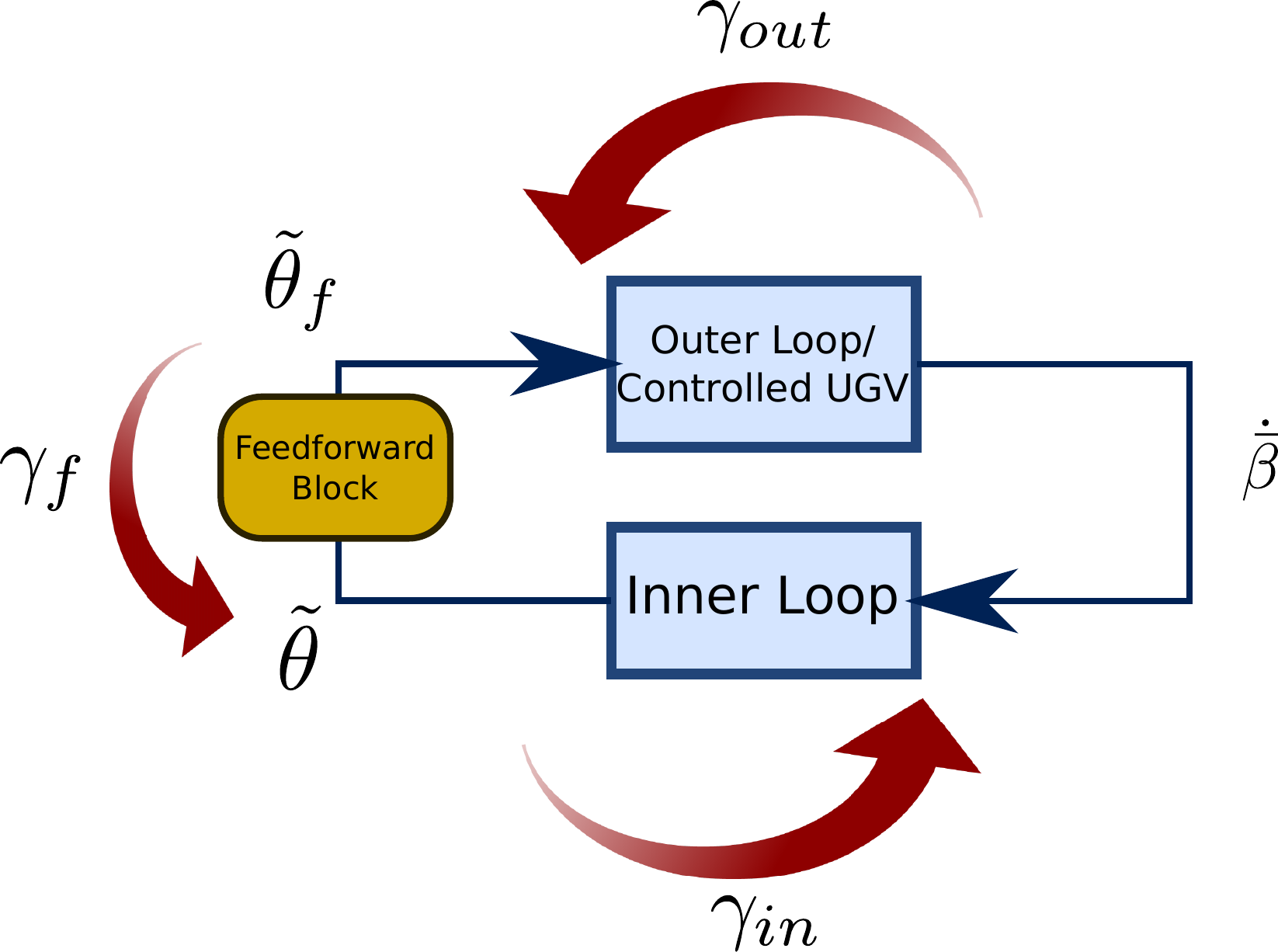}
		\caption{Interconnection of the system with the feedforward block.}\label{smallgainnew}}
	\end{figure}

	Note that (\ref{u1new}) is equivalent to (\ref{u3}), (\ref{ftexplicit})-(\ref{maptheta}) and (\ref{u2control}) with a feedforward term that possibly reduces the effect of the attitude error on the outer loop. 
	In particular, the control law (\ref{u1new}) is equivalent to a feedforward block, which modifies the actual attitude error $\tilde{\theta}$ to $\tilde{\theta}_f$ (see Fig. \ref{smallgainnew}), where $\tilde{\theta}_f$ is such that
	\begin{equation}\label{thetaf}
		\dfrac{f_t\sin(\bar{\theta}+\tilde{\theta}_f)}{\sin\bar{\theta}}=\sigma_{0,T_{max}L}\left(\dfrac{f_t}{\sin\theta}\right)\sin\theta.
	\end{equation}

	To prove asymptotic stability, it is enough to prove that the gain introduced by the block is $\gamma_{f}\leq1$ or, in other words, that the action is able to reduce the effect of $\tilde{\theta}$. Under the assumption that $f_t\leq T_{max}L$, there are three cases:
	\begin{enumerate}
		\item If $\text{sign}(\sin(\bar{\theta}+\tilde{\theta}))=-\text{sign}(\sin\bar{\theta})$, then $f_{new}=0$, which corresponds to the case where $\tilde{\theta}_f=-\bar{\theta}$ using the previous mapping (\ref{mapu1}) (see Eq. (\ref{fnewfold}) and (\ref{thetaf})). 
			Note that $$\text{sign}(\sin\theta)=-\text{sign}(\sin\bar{\theta})$$ implies that $|\tilde{\theta}|\geq|\bar{\theta}|$ in the first law. As a consequence, $|\tilde{\theta}_f|\leq|\tilde{\theta}|$.
		\item If $\dfrac{f_t}{\sin\theta}\geq T_{max}L$, then $f_{new}=T_{max}L\sin\theta$. It is worth to remark that
			\begin{equation}\label{casetwoproof}
				\begin{aligned}
					\dfrac{f_t\sin(\bar{\theta}+\tilde{\theta})}{\sin\bar{\theta}}&\leq \dfrac{f_t\sin(\bar{\theta}+\tilde{\theta}_f)}{\sin\bar{\theta}}\leq f_t.
				\end{aligned}
			\end{equation}
			The inequalities (\ref{casetwoproof}) are equivalent to
			\begin{equation}
				\begin{aligned}
					\left|\dfrac{\sin(\bar{\theta}+\tilde{\theta})}{\sin\bar{\theta}}\right|&\leq\left|\dfrac{\sin(\bar{\theta}+\tilde{\theta}_f)}{\sin\bar{\theta}}\right|\leq1.
				\end{aligned}
			\end{equation}
			As a consequence, $0\leq|\tilde{\theta}_f|\leq|\tilde{\theta}|$.
		\item In all the other cases, no saturation occurs and $f_{new}=f_t$. In view of (\ref{fnewfold}) and (\ref{thetaf}), this case is equivalent to the first control law (\ref{mapu1}) where $\tilde{\theta}_f=0$.
	\end{enumerate}

	As a result, since $|\tilde{\theta}_f|\leq|\tilde{\theta}|$, the gain $\gamma_{f}$ between $\tilde{\theta}$ and $\tilde{\theta}_f$ can be smaller than one and, therefore, all the stability results of Theorem \ref{smallgainproposition} apply.

\end{document}